\documentclass[shortAfour,sageh,times]{sagej}
\usepackage[perpage]{footmisc}

\usepackage{moreverb,url}

\usepackage[colorlinks,bookmarksopen,bookmarksnumbered,citecolor=red,urlcolor=red]{hyperref}
\usepackage{amssymb}
\usepackage{amsmath}
\usepackage{breqn}
\usepackage{array}
\usepackage{comment}
\usepackage{relsize}
\usepackage{natbib}
\usepackage{setspace}
\usepackage{orcidlink}
\usepackage{enumitem}
\setlist{nosep}
\graphicspath{{./images/}}
\usepackage{adjustbox}
\usepackage{algorithm}

\usepackage{booktabs}

\newcommand\BibTeX{{\rmfamily B\kern-.05em \textsc{i\kern-.025em b}\kern-.08em
T\kern-.1667em\lower.7ex\hbox{E}\kern-.125emX}}

\def\volumeyear{2025}

\begin{document}


\title{Online Anti-sexist Speech: Identifying Resistance to Gender Bias in Political Discourse}

\author{Aditi Dutta\affilnum{1,3} Susan Banducci\affilnum{2}} 


\affiliation{\affilnum{1}University of Exeter, UK
\affilnum{2}University of Birmingham, UK
\affilnum{3}The Alan Turing Institute, UK
}

\corrauth{Aditi Dutta, University of Exeter,
Centre for Computational Social Science,
Knightley Building,
Exeter, Devon,
EX4 4PD, UK.}

\email{ad882@exeter.ac.uk}

\begin{abstract}
Anti-sexist speech—public expressions that challenge or resist gendered abuse and sexism—plays a vital role in shaping democratic debate online. Yet automated content moderation systems, increasingly powered by large language models (LLMs), may struggle to distinguish such resistance from the sexism it opposes. This study examines how five LLMs classify sexist, anti-sexist, and neutral political tweets from the UK, focusing on high-salience trigger events involving female Members of Parliament in the year 2022. Our analysis show that models frequently misclassify anti-sexist speech as harmful, particularly during politically charged events where rhetorical styles of harm and resistance converge. These errors risk silencing those who challenge sexism, with disproportionate consequences for marginalised voices. We argue that moderation design must move beyond binary harmful/not-harmful schemas, integrate human-in-the-loop review during sensitive events, and explicitly include counter-speech in training data. By linking feminist scholarship, event-based analysis, and model evaluation, this work highlights the sociotechnical challenges of safeguarding resistance speech in digital political spaces.

\end{abstract}


\maketitle

\begin{spacing}{0.55}
\noindent\textcolor{red}{\small\textbf{Content warning:} This document studies contents that may be offensive or upsetting. It will have illustrative examples of sexist languages online.}
\end{spacing}

\section{Introduction}\label{sec:intro}
\begin{quote}
    ``Anti-sexist campaigns are necessarily complex, and feminists differ on what they see as the most effective way of dealing with elements which they consider to be discriminatory.''
    \begin{flushright}- \citep{Mills_2008}\end{flushright}
\end{quote}

Significant progress has been made in identifying and addressing online sexism, with numerous studies dedicated to improving its detection across digital platforms (e.g., \citet{butt_2021, guest-etal-2021, nozza_volpetti_2019}, \textit{inter alia}).  While these approaches show impressive performance, they fail to identify and capture all forms of sexism or misogyny -- especially overlooking the subtler forms of sexist discourse \citep{rodriguez_2020, rodriguez_2021}, and are often prone to erroneous classifications.  There is a need to move beyond binary classifications of sexist/not sexist text \citep{dutta_2024}. In this paper we we move beyond binary classifications to understand language that actively resists sexism - what we conceptualize as anti-sexism. What exactly constitutes anti-sexism, and where does it appear in online discourse?

From a computational perspective, anti-sexist speech presents a unique challenge: it often shares linguistic and emotional characteristics (particularly tone, vocabulary and phrasing) with the sexist content it opposes. This makes it hard for language models to tell the difference between what a statement says and what it really means. More broadly, anti-sexism exemplifies a class of expressions that NLP systems routinely struggle to capture: those that challenge harmful ideologies using emotionally or lexically charged language. These challenges raise fundamental questions about how we train, evaluate, and trust language models when applied to socially situated discourse.

In political discourse, sexism and anti-sexism can be difficult to distinguish because they often use overlapping language, even though their intentions and meanings diverge sharply. Political speech frequently adopts a critical or emotional tone, which can obscure the difference between communication that reinforces and communication that challenges gender bias.  The implications of this, as \citet{ozer_2023} find, is that partisan polarization exacerbates gender-based stereotypes. This points to the difficulty of separating political ideology from prejudice and discriminatory communication \citet{ziems-etal-2024, burnham_2024}. Thus, political texts targeting women can be mistaken as sexist even when they are meant to defend or support them, especially because political language often adopts a negative and conflictual tone. Consequently, anti-sexist responses may be misread by both humans and models as sexist due to linguistic similarity or affective tone.

This study investigates how anti-sexism manifests in political discourse on Twitter (now X), specifically in posts directed at female Members of Parliament (MPs) in the United Kingdom. We assess whether current large language models (LLMs) can accurately classify such content by combining conceptual framing, prompt-based evaluation, and uncertainty calibration.

In the following sections, we draw on social psychology and political communication literature to conceptualise anti-sexist speech an important element of gendered online discourse. At the same time, we detail how not accounting for anti-sexist speech can perhaps diminish the reliability of the classification of sexist speech. We then describe out methods for classifying anti-sexist, sexist and political speech. In particular, we develop and implement new methods for assessing the reliability of classification. Our study highlights how automated systems may inadvertently suppress resistance speech (i.e., anti-sexism), thus exacerbating the very harms they aim to mitigate.

\vspace{0.5em}
\noindent In sum we make \textbf{four} primary contributions:
\begin{enumerate}[leftmargin=*]
\item We conceptualize and operationalize anti-sexism in online political discourse, grounding it in feminist and psychological theories (\S\ref{sec:intro}, \ref{sec:annotation}, Table~\ref{tab:def_terms}).
\item We evaluate multiple LLMs’ classification performance using prompt engineering with varying levels of instruction (\S\ref{sec:quant_analysis}).
\item We measure model confidence and trustworthiness in prediction using entropy and perplexity (\S\ref{sec:validity}).
\item We analyze how instruction style and complexity influence model confidence and performance (\S\ref{sec:imp_inst}).
\end{enumerate}

\noindent Our findings show that failure to treat anti-sexist speech as a distinct communicative category leads to significant misclassification, especially false positives in sexism detection. While current LLMs struggle to reliably distinguish these categories, omitting anti-sexism risks reinforcing platform biases that suppress resistance speech. Accurately modeling anti-sexism is thus not only a technical challenge but a political imperative for equitable content moderation.

\section{Conceptual Framework: Anti-Sexism and Political Discourse}\label{sec:conceptual}

Sexism itself is traditionally understood as behavior that reinforces patriarchal norms, it has shifted from overt forms to more subtle or benevolent manifestations \citep{manne_2017, he_2024}. \citet{Lovenduski_2014} notes how these biases affect not only political participation but also policy-making. Social psychology has long recognized sexism as multidimensional, such as \citet{eagly1994people} describe how women may be viewed “more positively” than men in some domains (benevolent sexism), while \citet{glick_fiske_1996} identify hostile and benevolent sexism as coexisting belief systems. Recent work shows that hostile sexism is more easily identified, especially by men, while women are more attuned to subtler forms \citep{kirkman2020just}. However, computational systems often rely on lexical cues associated with overt sexism and fail to detect subtler, context-dependent forms of sexism \citep{dutta_2024}.
To define anti-sexism in the context of online discourse, we draw on both feminist theory and behavioral studies. At its core, anti-sexism refers to speech or action that actively resists systemic and institutional sexism, including the structures of male privilege and patriarchy \citep{finch2004you}. As with anti-racism, anti-sexism involves not just rejecting overt prejudice, but recognizing one’s own social positioning and taking steps to challenge inequity. Theories of anti-racism and bystander intervention provide a valuable lens for identifying anti-sexist communication online \citep{DAVIS2021110724}. In particular, research highlights the importance of willingness to intervene in the face of sexist harm—whether through perspective taking \citep{DAVIS2021110724}, bystander behavior \citep{mcmahon2014}, or psychological flexibility in responding to injustice \citep{davis_1980}.

Behavioral l research offers valuable tools for broadening our understanding of anti-sexism by focusing not only on individual attitudes but also on how people recognize and respond to gender-based harm. A particularly relevant dimension is bystander behavior—the willingness to notice, interpret, and act in the face of sexist speech or abuse \citep{mcmahon2014}. Scales such as the Bystander Attitude Scale \citep{banyard2007sexual, mcmahon2014} are especially useful in this context, as they capture the propensity to speak out against injustice, challenge harmful norms, and offer support to targets of discrimination. These behavioral indicators can help inform a working definition of anti-sexism in online settings, where intervention often takes the form of informal, spontaneous responses to misogynistic or sexist speech. This perspective moves us beyond an attitudinal approach and provides concrete linguistic and affective cues for computational models to detect. As Mills \citep{mills_2003, mills_2003a} notes, it is important to resist reducing anti-sexist expression to superficial politeness or so-called ‘political correctness’; rather, it is a form of everyday resistance that can be systematically identified and studied. 

Against this background, we argue that recognizing anti-sexism is essential to improving the classification of sexism. For instance, a user critiquing sexist speech may use similar linguistic structure as the original post, leading to misclassification.  As \citet{Mills_2008} and \citet{pazhoohi_2013} argue, anti-sexist interventions in language can challenge dominant norms—but only if they are recognized correctly. Online platforms are central arenas for discourse analysis and automated content moderation \citep{Song_2020}, yet current tools may reinforce biases if they overlook anti-sexist speech. 

The emergence of large language models (LLMs) offers new avenues to explore these phenomena. Their strong performance on linguistic and cognitive tasks \citep{niu_2024, Balkus_2023} has positioned them as tools not only for classification but also for social inquiry. Researchers are increasingly using LLMs in data annotation pipelines, leveraging prompting techniques to direct model behavior in zero- and few-shot settings \citep{zhang_2022, meng_2022, tan_2024}. Studies suggest that instruction diversity significantly affects model performance in annotation tasks \citep{li_2024}, and in some cases, LLMs outperform human annotators on nuanced classification problems \citep{Zhuo_2023, bang_2023, ziems-etal-2024, Burnham_2024a}. Prompt engineering in LLMs, i.e., optimizing the language in a prompt in order to elicit the best possible performance, has become a critical interface between domain expertise and model generalization \citep{zhou_2023}.

However, LLMs are not without limitations. Their fluency can mask unreliability: they are prone to hallucinations \citep{huang2023, Ji_2023, farquhar_2024}, inconsistency \citep{Elazar_2021}, and overconfidence \citep{chen_2024, Alkaissi_2023}, which raises questions about the trustworthiness of their predictions. Misclassifying anti-sexism as sexism not only distorts public discourse but may also penalize users who challenge abuse, undermining the same voices that offer to counter sexism. Therefore, beyond simple accuracy, we measure models’ confidence and uncertainty in their predictions \citep{dong_2024}, ensuring that we assess both what the models predict and how sure they are.

\section{Context \& Sampling Rationale}\label{sec:context_rationale}

We examine anti-sexist speech by focusing on tweets directed at political figures, particularly female Members of Parliament (MPs) in the United Kingdom (UK). Existing research has shown that women in politics are often subjected to biased, sexist, and discriminatory treatment in media and public discourse, with politics identified as one of the slowest-moving domains in terms of gender equality \citep{Trabelsi_2023}. In the UK specifically, female MPs have been disproportionately targeted by online abuse—hostility that, as \citet{Scott_2023} notes, has “shut many women’s voices out of politics.”

In this context, we center our analysis on online conversations involving UK female politicians to examine both the presence of sexism and its counterpoint—anti-sexism—in digital political discourse. Twitter (now X) has long served as a prominent platform for political communication, especially among UK MPs \citep{agarwal_2021}. Its public nature and widespread political use make it a strategic site for studying gendered discourse. As \citet{zeinert-etal-2021} argue, Twitter is also methodologically advantageous, offering reduced selection and label bias compared to other social media platforms.

\subsection*{Trigger Events}\label{sec:trigger_event}\hfill\break
Offline events such as elections, protests, or political controversies often correspond with increases in online hate speech, sometimes in ways that appear only loosely related to the original event \citep{lupu_2023}. Drawing inspiration from \citet{dutta_2024a} and \citet{Kalyanam_2016}, we define \textbf{trigger points} as events—whether political, controversial, or otherwise personally related to a public figure—that incite a surge of online activity centered around the individual(s) involved and that have the potential to provoke sexist commentary.

For this study, we identified eight such trigger events (see Table~\ref{tab:time}) and collected Twitter conversations surrounding them. These events were categorized into two types---sexist and political, based on which events generated higher concentrations of targeted online activity. While both types involve female politicians and carry the potential to prompt sexist reactions, our aim is not to establish a causal link between the events and sexism. Instead, we treat them as high-salience entry points for sampling online discourse, thereby increasing the likelihood of capturing both sexist and anti-sexist speech. Although our time frame is limited to a few days around each trigger event, this design allows us to examine whether event type and engagement level (as filtered through engagement metrics) influence the prevalence of sexist and anti-sexist conversations online.


\begin{table}[ht!]
\setlength{\tabcolsep}{2pt}
\footnotesize
  \begin{tabular}{p{1cm}p{1.5cm}p{13.5cm}}
    \hline 
    \textbf{Type} & \textbf{Date} & \textbf{Event} \\[0.5mm]
    \hline
    {Political} & \textit{February} & {On 24 February 2022, Diane Abbott was one of eleven Labour MPs who signed a statement by the Stop the War Coalition criticizing the UK government.}\\
    \hline
    {Political} & \textit{March} & {On 15 March 2022, Priti Patel was the victim of a prank video call by Russian comedians Vovan and Lexus who were accused by Britain of working for Russia.}\\
    \hline
    {Sexist} & \textit{April} & {On 24 April 2022, Angela Rayner was the subject of a report in The Mail on Sunday, by Glen Owen, in which it was alleged that she had tried to distract Boris Johnson in the Commons by crossing and uncrossing her legs in a similar manner to Sharon Stone in a scene from the 1992 film Basic Instinct.} \\
    \hline
    {Political} & \textit{May} & {On 9th May, Rayner said she would resign if she received a fixed penalty notice for breaching COVID-19 regulations while campaigning during the run-up to the Hartlepool by-election and local elections the previous year.}\\
    \hline
    {Political} & \textit{May} & {On 11 May 2022 Jess Phillips narrowly avoided being referred to the Parliamentary Committee on Standards after being investigated by the Commissioner for Standards for repeatedly failing to register interests within the required timescale.}\\
    \hline
    {Political} & \textit{September, October} & {On 5 September 2022, in anticipation of the appointment of Liz Truss as Prime Minister, Priti Patel tendered her resignation as Home Secretary which was effective from 6 September. Liz Truss was appointed as Prime Minister by Queen Elizabeth II at Balmoral Castle on 6th September. She was succeeded by Rishi Sunak as leader of the Conservative Party on 24 October.}\\
    \hline
    {Political} & \textit{October} & {On 25 October, Suella Braverman was reappointed as the Home Secretary by Prime Minister Rishi Sunak upon the formation of the Sunak ministry.}\\
    \hline
    \end{tabular}
    \captionsetup{singlelinecheck=off}
    \caption{\small This table documents the different types of trigger events we took into consideration, along with their respective months. The controversies centering around the target female politicians are also mentioned. Though the conversations were mostly centered around the said politicians, mentions of the other politicians we considered were also found in our data. Conversations during these events were further selected based on the metrics of engagement, as stated in $\mathsection$\ref{sec:data_prep}.}
  \label{tab:time}
\end{table}



\section{Data and Annotation}
\subsection{Data Collection}\hfill\break

We collected data from X (formerly Twitter) via academic API access for the year 2022, a period of heightened political and economic turbulence in the United Kingdom—including three Prime Ministers, the death of Queen Elizabeth II, and a worsening economic crisis \citep{middleton_2023}. This period was selected to capture a wide range of politically charged and controversial events, increasing the likelihood of encountering both sexist and anti-sexist discourse.
We began by identifying 38 UK female Members of Parliament (MPs) based on their political visibility and consistent public engagement on X. Using a set of manually selected trigger events (see $\mathsection$\ref{sec:trigger_event}), we collected original, reply, and quoted tweets related to these MPs\footnote{Please check the different types of tweets expected from the Twitter platform: \url{https://developer.twitter.com/en/docs/tutorials/determining-tweet-types}}. We excluded tweets authored by the MPs themselves and omitted retweets, as they typically reflect diffusion rather than original commentary. To prioritize high-impact content for annotation, we ranked tweets using engagement metadata (likes, replies, quotes, retweets) and removed duplicates.

\subsection{Data Preparation}\label{sec:data_prep}\hfill\break
To prepare the data for annotation and model analysis, we applied a structured cleaning pipeline designed to reduce noise and improve interpretability. First, we removed empty entries, excess whitespace, duplicate tweets, and non-English texts. Tweets composed entirely of emojis or URLs were also excluded, as such content is often ambiguous and difficult for both human annotators and automated classifiers to interpret reliably. We additionally filtered out institutional or news-style content using a keyword-based approach (e.g., ``BREAKING NEWS'', ``HEADLINES''), in order to focus exclusively on public discourse from individual users rather than formal media narratives.\footnote{See $\mathsection$\ref{sec:filtering} for a full filtering pipeline.} Emoticons were standardized, and contractions were expanded (e.g., ``don’t'' to ``do not'') to enhance text clarity. After cleaning, we prioritized tweets with higher public engagement by sorting them based on metadata such as the number of likes, replies, retweets, and quotes. This allowed us to focus annotation efforts on posts that attracted greater public attention during the identified trigger events. Duplicate entries across these sampling windows were removed to ensure data quality and prevent interpretive redundancy.

\subsection{Annotation Process}\label{sec:annotation} 

Our annotation process was conducted in two phases. In the initial pilot phase, two annotators labeled a total of 833 tweets, with each tweet annotated by a single individual. This phase allowed us to test label clarity, refine definitions, and identify sources of ambiguity. In the second phase, we engaged nine annotators—both male and female—who were experts in gender studies and familiar with UK political discourse. The remaining 510 tweets were labeled by three annotators each, enabling us to capture a broader range of interpretations.

To support consistent annotation, all annotators were provided with a detailed guidelines document that included our research objectives and category definitions—\textit{sexist}, \textit{anti-sexist}, and \textit{neither}—as well as multiple labeled examples per class\footnote{See $\mathsection$\ref{appendix:labels} for full class definitions and examples.}. We also held review sessions to clarify uncertainties and ensure mutual understanding. These three categories, grounded in feminist and behavioral theory (see $\mathsection$\ref{sec:conceptual}), were also used to prompt the large language models (LLMs), allowing for direct comparison between human and machine judgment.

Rather than adopting a majority voting scheme, which prior work has shown to require large annotation pools and may still fail to reflect epistemic diversity \citep{sheng_2008, Hube_2019, wallace_2022, fleisig_2024, basile_2021}, we opted for a minority voting strategy.\footnote{In this scheme, the label receiving the fewest votes is selected. See Section~\ref{sec:minority-label} for details.} This approach acknowledges the subjective nature of labeling sexism and anti-sexism, and preserves interpretive disagreement as a valuable signal rather than treating it as noise. As others have argued, disagreement in subjective tasks does not necessarily imply error \citep{Zheng_2017, fleisig_2024, basile_2021}.
Each tweet was therefore labeled by three annotators using the three-class scheme, with the final label determined by selecting the minority vote. In the rare case of a three-way tie, we applied a fixed resolution strategy: \textit{sexist} was prioritized over \textit{neither}, which was prioritized over \textit{anti-sexist}. Notably, no cases involved direct conflict between the \textit{sexist} and \textit{anti-sexist} labels, suggesting that most disagreement occurred between polar and neutral interpretations.

We defined three primary categories for annotation: \textit{sexist}, \textit{anti-sexist}, and \textit{neither}, drawing on feminist and behavioral theories of online speech (see $\mathsection$\ref{sec:conceptual}). 
The same three categories were used to prompt large language models (LLMs) for classification. Prompts were designed to test the influence of instruction style, specificity, and reasoning format on model output. 


\section{Methodology}\label{sec:methodology}


\subsection{LLMs for Prompt Engineering}\label{sec:prompt_engineering}
\hfill\break
Several LLMs have shown promise in sexism detection, but for this study, we took the top five open-source LLMs which have shown high-performance results in some of the other existing sexism datasets. The \textbf{five} LLMs in question are: Gemma-7B instruct version\footnote{\tiny \url{https://huggingface.co/google/gemma-7b-it}} \citep{gemma}, 
Flan-T5-xl\footnote{\tiny \url{https://huggingface.co/google/flan-t5-xl}} \citep{flant5}, Mistral-7B instruct version\footnote{\tiny \url{https://huggingface.co/mistralai/Mistral-7B-Instruct-v0.2}} \citep{mistral7b}, Qwen2.5-0.5B-Instruct\footnote{\tiny\url{https://huggingface.co/Qwen/Qwen2.5-0.5B-Instruct}} \citep{qwen2.5} and Llama 3.1-8b\footnote{\tiny\url{https://llama.meta.com/}} \citep{llama}.
Since LLMs are trained on huge amount of public data, making them capable of understanding and generating natural language and other types of content to perform a wide range of tasks \citep{Ibm_2024}, they are widely being adapted by researchers for annotations due to its cost-efficiency an ease of implementation. Not only that, they also provide a valuable resource of labeled data for NLP models in different stages, presenting solutions to challenges like data scarcity, enhancing annotation quality and process efficiency \citep{tan_2024}.
\citet{wang_2023} proved that LLMs can effectively be used as zero-shot text classifiers since it consistently outperformed many traditional NLP algorithms in some datasets containing varied classification tasks. Inspired from them, we use LLMs-as-classifiers using prompt engineering, where it employs a single-step prompt-based approach called \emph{zero-shot prompting}. The streamlined approach is intended to leverage the LLMs' generative capabilities to directly produce a specific classification label, when directed with varied levels of instructions (also referred in this work as \emph{prompt categories}). The classification process is formalized as: $LLM\ response \ (prompt) \rightarrow classification\ output$. By using this method, it simplified the classification process and eliminated the need for intermediate steps like feature extraction or explicit verbalizer mapping \citep{wang_2023}. \citet{ziems-etal-2024} have shown to offer a data-efficient alternative to fine-tuned models in capturing political opinions (among others), through efficient prompt tuning by reducing the computational costs, which motivates us more to use the same approach in our work.
\hfill\break

\subsubsection*{Prompt Stability or Paraphrase Robustness} 
\hfill\break
Responses from LLMs are usually susceptible to the influence of the choice of the prompts \citep{griffin2023susceptibility}, where minimal semantics-preserving prompt template paraphrases are seen to substantially affect the output result where the differences in the response are larger than expected \citep{röttger2024}. 
Studies have explored multiple ways to improve LLMs' response to prompts. 
Similar to \citep{dutta_2024a}, we use several prompt settings that measure the variation of the performances among several prompt structures. We started off with simple examples that the LLMs had to validate as either of the classes, gradually progressing towards difficult instances. To evaluate the prompt effectiveness based on the output quality of the respective LLM, we used four kinds of prompt evaluation metrics\footnote{\small\textit{"How to measure the quality of LLMs, prompts and outputs"}\citep{Cheung_2024}}: (i) grounding (the authoritative basis of the LLM output, determined by comparing it against some ground truths in a specific domain), (ii) relevance (how relevant the LLM’s response is to a given user’s query), (iii) efficiency (the speed and computing consumption of the LLM to produce the output.), and (iv) hallucinations (looking at LLM hallucinations with regard to retrieved context). For most of the models, all of these four metrics gave positive outcome within a few trials (except Mistral). Since \citet{gonen-etal-2023} suggests that prompts with lower perplexity score (concept explained in $\mathsection$\ref{sec:confidence_estimation}) performs better task, we further checked for the perplexity in the generated text in the multiple prompt templates, and chose to go with the prompt which showed the least perplexity.
We develop a general template of prompts (see Table \ref{tab:prompt categories} and \ref{tab:def_terms}), which we re-used across all the LLM prompt styles, adjusting according to the required context length for each LLMs, as per their prompt strategies. 


\subsubsection{Notations}\label{sec:notations}\quad

\noindent Given a dataset $\mathcal{D} := [(x_1, a_1), \dots, (x_N, a_N)] \subset \mathcal{X} \times \mathcal{Y}$ consists of $N$ tuple instances containing a text $x$, and a ground truth (or annotated label) $a$, where $\mathcal{X}$ is the instance space containing the text and $\mathcal{Y}$ is the set of outcomes. The dataset examples $(t_i, a_i)$ are assumed to be independent and identically distributed (i.i.d.) according to some unknown probability measure $p(.,.)$ on $\mathcal{X} \times \mathcal{Y}$.
For an $i$-th instance $x_i$ is prompted, it will generates an output $f:x_i \mapsto f(x_i)$, which is denoted by $[y_i^1, \dots, y_i^l, \dots, y_i^L] = y_i \in \mathcal{Y}$ when consisted of a sequence of $L$ tokens given an input $x_i$. Since ours is a classification task, $l$ would ideally be one token representing one category of the label class, unless the output gives back a sequence of length $L$. An input $x_i$ would contain linguistic cues depending in the prompt $P$ based on the prompt category $r$, and a question for the classification task (same across all prompt categories) $q$, hence can be denoted as $x_{i;r,q} = [P;r_i,q_i]$ and will give an output $y_{i;r,q}$.

\subsection{Estimating Model Uncertainty and Interpretability}\label{sec:interpretability}

To better understand how confident models are in their predictions (especially in ambiguous or contested cases), we assessed model uncertainty using standard probabilistic and divergence-based techniques. Specifically, we examined:

\begin{itemize}
    \item \textbf{Predictive entropy} to evaluate how certain models are about their top predictions across sampled completions;
    \item \textbf{Perplexity} to estimate the language model’s confidence in token-level generation; and
    \item \textbf{Jensen–Shannon Divergence} to capture how much the model’s label probabilities differ across alternative prompts.
\end{itemize}

These measures help reveal whether the models are not only accurate but also calibrated, i.e., whether they ``know when they don’t know.'' Such insights are particularly important when dealing with complex, subjective categories like sexism and anti-sexism, where false confidence can carry real-world risks.

In our findings (detailed in $\mathsection$\ref{sec:results}), we observe that models often appear highly confident even when misclassifying anti-sexist speech. This suggests that commonly used uncertainty metrics may not be reliable indicators of model reliability in politically or socially sensitive contexts.
For full mathematical details and implementation specifics, see $\mathsection$\ref{sec:uncertainty}.


\section{Results}\label{sec:results}


\subsection{Performance Evaluation}\label{sec:quant_analysis}

\begin{table*}[ht!]
\centering
\setlength{\tabcolsep}{2pt}
\renewcommand{\arraystretch}{1.2}
\begin{adjustbox}{width=\textwidth,keepaspectratio}

  \begin{tabular}{|l|c|c|c|c|c|c|c|c|c|c|c|c|c|c|c|c|c|c|c|c|}
    \hline
    \multicolumn{1}{|c|}{\small\textbf{Prompt}} & \multicolumn{4}{c|}{\small\textbf{Mistral}} & \multicolumn{4}{c|}{\small\textbf{Llama}} & \multicolumn{4}{c|}{\small\textbf{FlanT5}} & \multicolumn{4}{c|}{\small\textbf{Qwen}} & \multicolumn{4}{c|}{\small\textbf{Gemma}}\\
    \cline{2-21}
    \multicolumn{1}{|c|}{\small\textbf{Type}} & \multicolumn{1}{c|}{R} & \multicolumn{1}{c|}{P} & \multicolumn{1}{c|}{F1} & \multicolumn{1}{c|}{A} & \multicolumn{1}{c|}{R} & \multicolumn{1}{c|}{P} & \multicolumn{1}{c|}{F1} & \multicolumn{1}{c|}{A} & \multicolumn{1}{c|}{R} & \multicolumn{1}{c|}{P} & \multicolumn{1}{c|}{F1} & \multicolumn{1}{c|}{A}
    & \multicolumn{1}{c|}{R} & \multicolumn{1}{c|}{P} & \multicolumn{1}{c|}{F1} & \multicolumn{1}{c|}{A} & \multicolumn{1}{c|}{R} & \multicolumn{1}{c|}{P} & \multicolumn{1}{c|}{F} & \multicolumn{1}{c|}{A}\\
    \hline
    {Roleplay} & 42.83 & \textbf{\underline{51.77}} & 44.93 & \textbf{\underline{82.60}} & 44.07 & 40.53 & 39.66 & 67.86 & \textbf{\underline{54.53}} & 50.06 & \textbf{\underline{51.11}} & 76.97 & 16.78 & 20.58 & 18.48 & 56.12 & 40.96 & 36.70 & 19.69 & 21.20\\
    {Content} & 46.04 & \textbf{\underline{64.45}} & \textbf{\underline{48.77}} & \textbf{\underline{84.44}} & \textbf{\underline{50.13}} & 43.55 & 33.40 & 38.55 & 46.46 & 51.42 & 47.08 & 79.28 & 0.39 & 16.67 & 0.75 & 1.61 & 45.41 & 41.10 & 30.98 & 38.62\\
    {Zero-shot} & \textbf{\underline{52.19}} & \textbf{\underline{56.50}} & \textbf{\underline{53.59}} & \textbf{\underline{82.02}} & 42.94 & 33.26 & 33.14 & 59.49 & 44.76 & 51.11 & 41.45 & 77.69 & 14.44 & 16.55 & 15.43 & 60.39 & 45.58 & 40.03 & 23.72 & 25.81\\
    {Few-shot} & 53.12 & \textbf{\underline{48.08}}  & 47.95 & 71.24 & \textbf{\underline{56.26}} & 43.14 & 40.55 & 53.14 & 47.00 & \textbf{\underline{55.97}} & 46.34 & \textbf{\underline{80.42}} & 11.70 & 16.34 & 13.64 & 48.92 & 48.52 & 39.62 & 32.26 & 40.53\\
    
    \hline
    \end{tabular}
  \end{adjustbox}
  \caption{\small This table documents the performance metrics we used in our study: macro-Recall \textbf{(R)}, macro-Precision \textbf{(P)}, macro-F1 \textbf{(F1)} and accuracy \textbf{(A)} scores for each models and their prompt categories. All measures are recorded in percentage(\%). The best scores of each metrics are highlighted in bold and underlined.}
  \label{tab:performance}
\end{table*}

As we see from Table \ref{tab:performance}, the performance across the range of models and the prompt types were not promising. Most of the models seem to under-perform in detecting sexism and anti-sexism against other texts, in the political discourse through prompt engineering. The disparity in performance between LLMs can be attributed to the specific tuning conducted to optimize their pre-trained versions for chat compatibility \citep{kumarage2024harnessing}. We report our experimental results using macro-averaged scores of multiple classification evaluation metrics (accuracy, recall, precision, F1-score), given the imbalance in the dataset. Ideally, the recall score is favourable over other evaluation metrics since recall is the measure of the ability of a model to define the true positive sexist and anti-sexist speech. Having a lower recall would suggest that there are larger linguistic patterns that the model would not be able to detect \citep{warner2012detecting}. Yet some studies, such as \citet{ayo_2020} say that precision too is useful to ensure hate speech projections are as precise as possible. This too is necessary for this research, as the main purpose is to differentiate between sexist and anti-sexist texts. We see that Mistral gives the best overall performance, with having higher precision value for all the prompt types, among the models. Llama3 and FlanT5 too give comparable performances, yet Gemma and (especially) Qwen give the worst performances. The experimental results indicate that adding more instructions to the model can enhance model performance in some scenarios, but not in all. 
It is noteworthy that in many instances, such as in Mistral and FlanT5, few-shot in-context learning performs worse than zero-shot performance (e.g., Mistral), and even as compared to roleplay (zero-shot) prompting (e.g., FlanT5), as indicated by their macro-F1 score. Most significantly, the models usually seem to encounter the most failures on the content (zero-shot) prompting type set despite providing the models with content information, to support their prediction in political discourse.

\begin{figure}[htb!]
    \centering   
    \begin{minipage}[b]{\linewidth}
    \centering
    \includegraphics[width=0.9\linewidth]{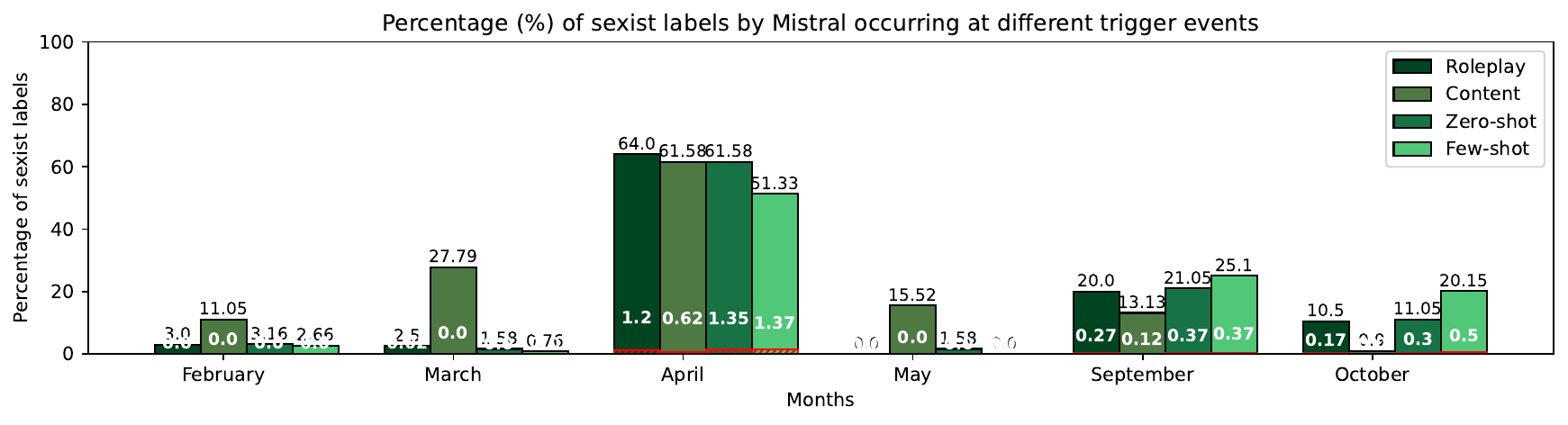}
    \captionof{figure}*{\centering\small(A): Sexism}
  \end{minipage}\hfill
    \begin{minipage}[b]{\linewidth}
    \centering
    \includegraphics[width=0.9\linewidth]{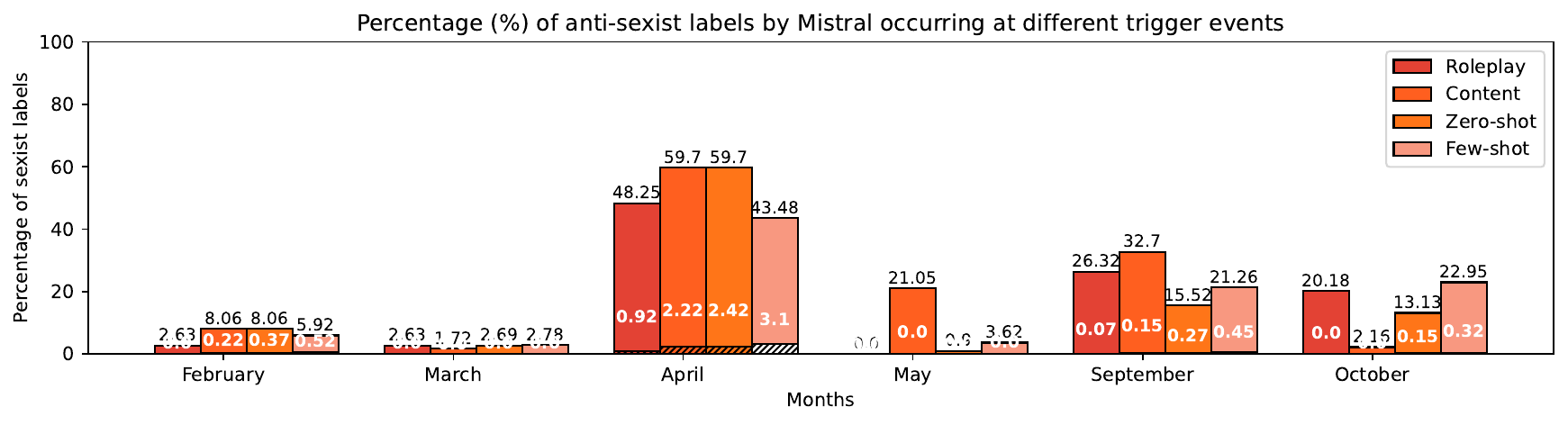}
    \captionof{figure}*{\centering\small(B): Anti-sexism}
  \end{minipage}\hfill
  \caption{\small This figure shows the proportion of tweets predicted as sexist (A) and anti-sexist (B) by the Mistral model across six major trigger-event months in 2022. Bars are grouped by prompt type: roleplay, content, zero-shot, and few-shot. Shaded segments indicate the proportion of correctly classified instances. While Mistral frequently overpredicts both categories (particularly around high-salience events like April) its accuracy remains low, especially for anti-sexist speech. These patterns reveal the model’s difficulty in distinguishing between harmful and resistant language when both share similar tone and phrasing.}
\label{fig:sexism_antisexism_prop}
\end{figure} 

To further delve into the precision and recall of the models specifically for sexism and anti-sexism, we record the proportion of these categories as predicted by all the models (except Qwen, owing to its bad performance) into bar-plots over each month. Figure \ref{fig:sexism_antisexism_prop} compares how models predicted instances of both sexist (A) and anti-sexist (B) speech across trigger events, disaggregated by prompt type. Full model-by-model breakdowns are available in Appendix \ref{appendix:perp_sex_anti}.
In each figure, each bar plot for a certain month indicates different prompt types, with the shaded portions at the bottom representing the proportion of sexist and anti-sexist that were correctly predicted. As we see in both the figures, the models tend to over-estimate the presence of both sexist and anti-sexist instances, especially at the event of a sexist trigger event (in the month of April, check Table \ref{tab:time}). Though the proportions of correctly predicted sexist and anti-sexist texts are small, we see a general rise in proportions in both the categories, when more instructions (and examples) are added. 

This shows slight contradiction to the scenario indicated by the performance metrics, yet leaves us to believe that maybe adding instructions does make the model more confident in (correctly) recognizing a text as sexist or anti-sexist, which is favourable for sexism identification tasks.

\begin{figure}[htb!]
    \centering   
    \begin{minipage}[b]{.9\linewidth}
    \centering
    \includegraphics[width=\linewidth, keepaspectratio]{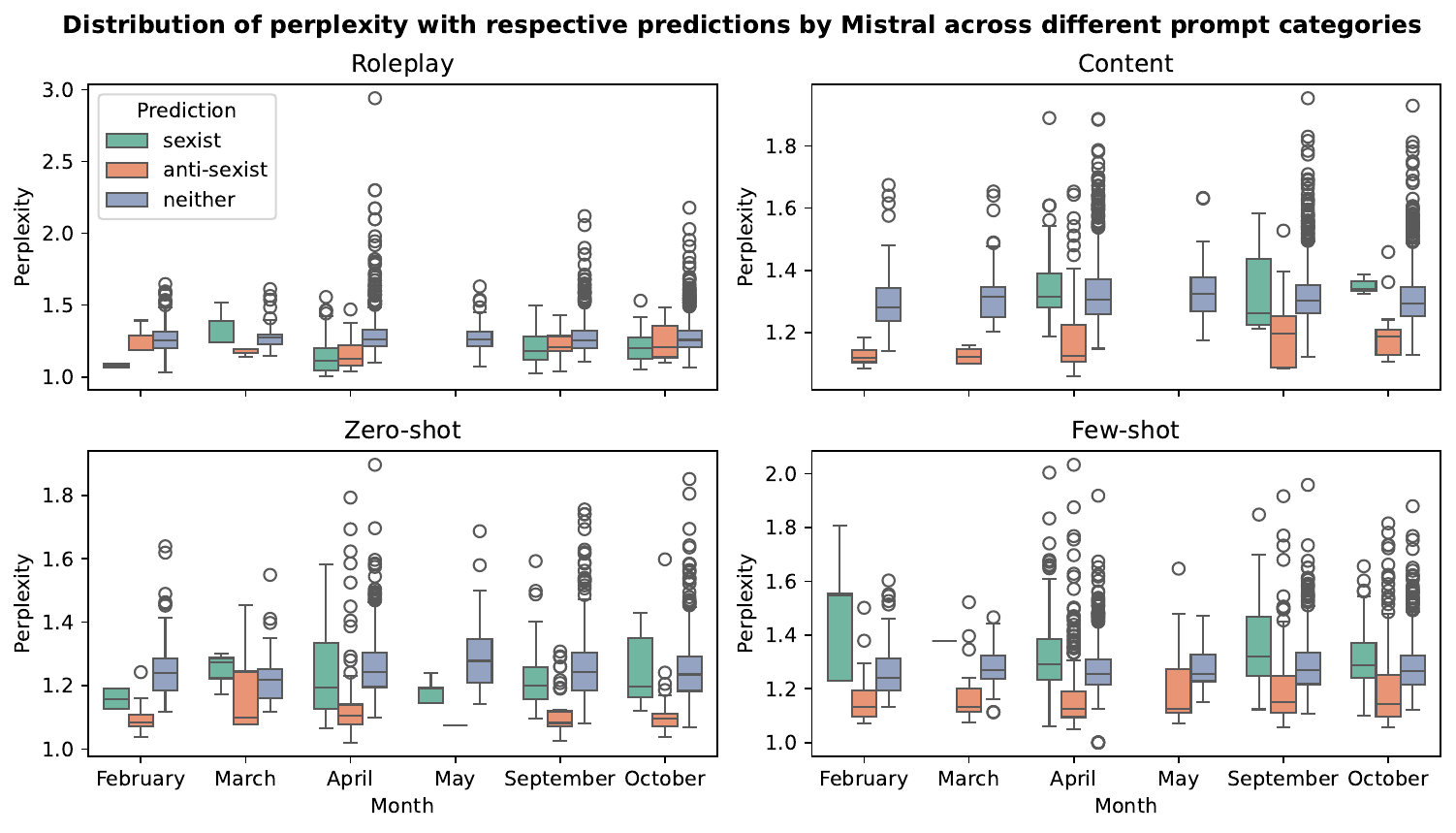}
    \captionof{figure}*{\centering\small(A)}
  \end{minipage}\hfill
  \break
  \begin{minipage}[b]{.7\linewidth}
    \centering
    \includegraphics[width=\linewidth, keepaspectratio]{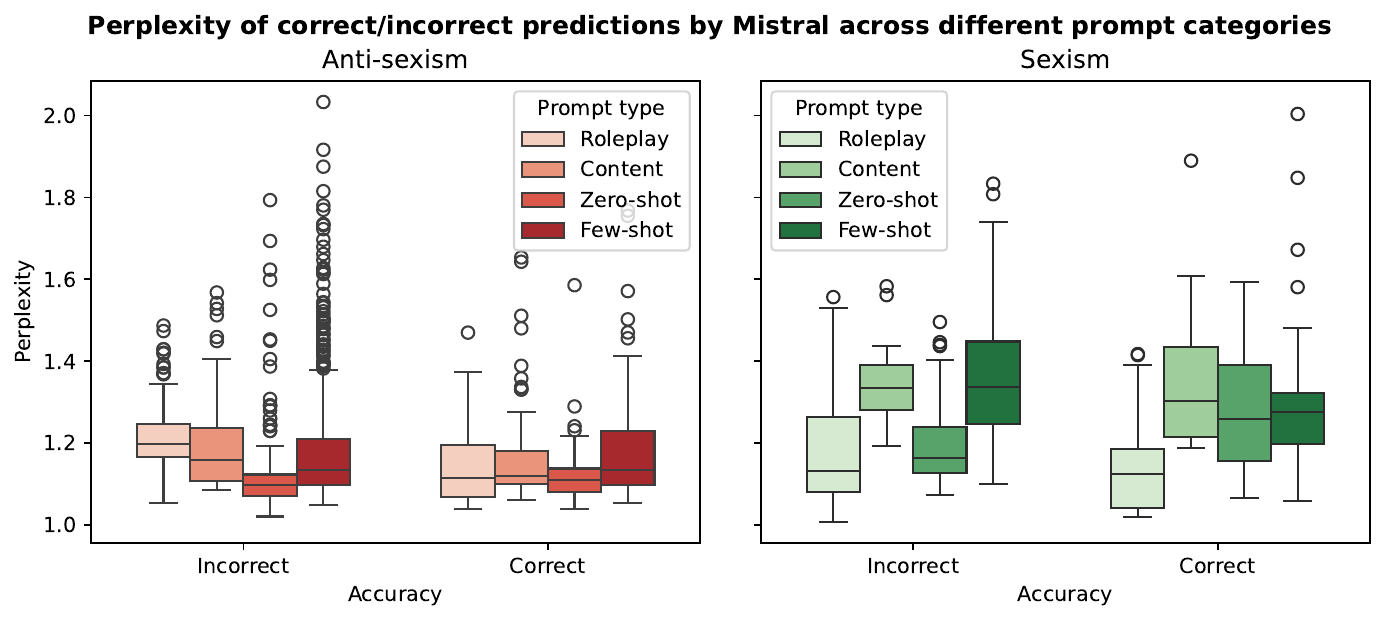}
    \captionof{figure}*{\centering\small(B)}
  \end{minipage}\hfill
  \caption{\small The figures demonstrate boxplots representing the distribution of the perplexity score for Mistral. In figure (A), the perplexity is shown across each class category across the months, while in figure (B), the perplexity is shown across the key classes- \emph{sexist} and \emph{anti-sexist} for each prompt type, differentiating based on the accurate predictions by the model. Each bar shows the proportion of tweets labeled as sexist, anti-sexist, or neither by the four prompt categories. Across almost all categories, sexist labels are overpredicted, while anti-sexist speech is under-recognized. This misclassification pattern suggests that LLMs struggle to differentiate between harmful and resistant gendered speech.}
\label{fig:mistral_perp}
\end{figure}

\subsection{Models are Confident, But are their Responses Valid?}\label{sec:validity}

\begin{table}[ht!]
\setlength{\tabcolsep}{2pt}
\footnotesize
\centering
\setlength{\tabcolsep}{3pt}
\renewcommand{\arraystretch}{1.1}
\begin{adjustbox}{width=0.96\textwidth,keepaspectratio}
        \begin{tabular}{c|c|c|c|c|c|c|c|c|c|c|c|c|c|c|c|c}
            \hline
            {} & \multicolumn{4}{|c|}{Ambiguous} & \multicolumn{4}{c|}{Non-ambiguous } & \multicolumn{4}{|c|}{Conflict} & \multicolumn{4}{c}{Confident}\\
            {Models} & \multicolumn{4}{|c|}{(when annotators disagreed)} & \multicolumn{4}{c|}{(full consensus)} & \multicolumn{4}{|c|}{(all predictions not same)} & \multicolumn{4}{c}{(all predictions same)}\\[0.1mm]
            \cmidrule{2-17}
             {} & {R} & {C} & {Z} & {F} & {R} & {C} & {Z} & {F} & {R} & {C} & {Z} & {F} & {R} & {C} & {Z} & {F}\\
             \hline
            {Mistral} & 1.292 & 1.306 & 1.251 & 1.258 & 1.269 & 1.297 & 1.237 & 1.261 &1.326 & 1.383 & 1.337 & 1.365 & 1.271 & 1.309 & 1.237 & 1.251\\
            {Llama} & 1.305 & 1.156 & 1.331 & 1.209 & 1.314 & 1.187 & 1.363 & 1.235 & 1.489 & 1.324 & 1.505 & 1.341 & 1.279 & 1.125 & 1.296 & 1.157\\
            {FlanT5} & 1.078 & 1.022 & 1.020 & 1.006 & 1.049 & 1.012 & 1.015 & 1.010 & 1.240 & 1.218 & 1.205 & 1.209 & 1.035 & 1.027 & 1.019 & 1.023\\
            {Qwen} & 1.385 & 1.161 & 1.657 & 1.473 & 1.397 & 1.160 & 1.664 & 1.417 & 1.469 & 2.122 & 1.894 & 1.681 & 1.320 & 1.165 & 1.513 & 1.393\\
            {Gemma} & 25.029 & 5.399 & 4.761 & 5.091 & 9.112 & 9.293 & 9.458 & 16.999 & 16.208 & 66.788 & 28.497 & 34.604 & 8.866 & 5.915 & 7.153 & 8.799\\
            \hline
    \end{tabular}
  \end{adjustbox}  
     \caption{\small This table demonstrates the mean perplexity of the respective models, in two situations: i) human (dis/)agreement [ambiguous/non-ambiguous], and ii) model (dis/)agreement [conflict/confident]. The model disagreement is assumed when the predictions generated for a particular input varies through multiple sampling, thus indicating confusion in the model. While model disagreement was collected for the whole dataset, human (or annotator) disagreement was collected only for the instances from phase 2 annotation task ($n=510$). The double-digit perplexity scores are due to some outliers which ranged further than 50, as shown in later boxplots.}
     \label{tab:avg_perp}
  \break
\end{table}

In the previous sub-section, we learnt that models were getting more confident in recognizing sexist and anti-sexist instances, but are their predictions reliable? In this following section, we explore the confidence estimation of the models using two metrics -- perplexity and predictive entropy scores. 

\subsubsection*{Exploring Model Predictions Around Key Trigger Events}\label{sec:pred_trigger}\quad

Model predictions varied considerably across the six trigger-event months, reflecting differences in both discourse tone and event salience. In February’s parliamentary policy debate, sexist and anti-sexist classifications were relatively low (below 10\%), consistent with the event’s more technical and procedural nature. During this month, model outputs aligned more closely with annotation consensus for anti-sexist than sexist, suggesting that less emotive political discourse reduces classification confusion for anti-sexism.
In April, the sexist controversy surrounding MP Angela Rayner generated heightened political polarization and gender-focused commentary. Mistral, along with other models, classified a substantial share of tweets as sexist, and accuracy seem higher than the political trigger events.
Both September and October’s leadership transition led to a moderate rise in sexist predictions ($\approx$20\%) as compared to other political trigger events, while anti-sexist detection surprisingly also increased. Yet accuracy dropped as compared to the sexist trigger event, implying that emotionally charged and adversarial discourse amplifies misclassification risk. The event’s focus on political power struggles rather than direct gendered attacks may explain the muted recognition of both sexism and anti-sexism.
Overall, these event-linked trends suggest that models are particularly vulnerable to misclassification during high-salience political moments which does not involve gender controversy, where the linguistic overlap between harmful and resistant speech is greatest.

\subsubsection*{Exploring Perplexity scores}\label{sec:perplexity_score}\quad

Like we discussed before, perplexity is used NLP tasks for evaluating confidence of a LLM, in a way capturing the degree of uncertainty a model has in predicting a generated sequence (here, classification category). A lower perplexity score is generally better, as it indicates more confidence of the models (and accuracy in its predictions in some cases). The score might vary depending on the model type, data and task, even though it does not have an upper bound. Since a classification task would not (ideally) generate a lot of text, the perplexity score is expected to be low -- as low as less than 5. While we concluded about the accuracy in the previous section ($\mathsection$\ref{tab:performance}), we want to test how confident the model is while classifying the text, and also identify if any disagreement in judgment (by both human and model) might factor in the perplexity measure. In Table \ref{tab:avg_perp}, we explore the perplexity scores of the models across the prompt types, with different conditions of disagreement in judgments. We find that model perplexity increases with both human annotation disagreement and model prediction inconsistency, confirming that perplexity effectively reflects uncertainty. FlanT5 consistently exhibits the lowest perplexity, indicating strong confidence and stability, while Gemma shows extremely high perplexity (especially during model disagreement) signaling unreliable performance. Differences between prompt types are present but not uniform across models; however, few-shot prompting generally reduces uncertainty. Overall, models are more affected by their own internal disagreement than by human label ambiguity, highlighting limitations in capturing nuanced social language like one would expect in anti-sexist contents.

We also plot the spread or distribution of perplexity in each model, based on the class category and accuracy (for sexist and anti-sexist predictions only). Detailed analyses of model accuracy and confidence distributions are available in Figure \ref{fig:mistral_perp} and Appendix \ref{appendix:pred_label_mod}.
In general, we see that the distribution is more symmetric for class category \emph{neither}, while sexist and anti-sexist texts are usually skewed. Furthermore, anti-sexist texts demonstrate lesser variation to perplexity than sexist texts. This could be a result of adding further information on both anti-sexism and sexism to the model. While the model is introduced to the newer concept of anti-sexism through the instructions, adding instructions on sexism may tend to confuse the model further, if the information are very specific for the dataset as they may tend to override their prior knowledge (parametric-based) and biases from their pre-training stage. Furthermore, we do not see a huge difference between the perplexity scores in the model between the ambiguous and non-ambiguous, and conflicted and confident texts. The confident texts (understandably) tend to have a lower perplexity scores than the conflicted ones, owing to the different category selecting during classification. But the ambiguous text show the same perplexity as the non-ambiguous ones. In summary, while all models differ in the extent to which they favor each category, they consistently overidentify sexist content and fail to reliably detect anti-sexist speech. This trend supports our central claim that models often conflate counter-speech with toxicity, especially in politically charged contexts.


\subsubsection*{Exploring Predictive Entropy scores}\label{sec:pred_ent_score}\quad

Another confidence metric, the PE measures the uncertainty in the LLM's prediction. Similar to perplexity, a lower predictive entropy would be preferable, but unlike the previous one, we calculate PE on the probability distribution of the model over possible outcomes.

\begin{table}[ht!]
\setlength{\tabcolsep}{2pt}
\footnotesize
\centering
\setlength{\tabcolsep}{3pt}
\renewcommand{\arraystretch}{1}
\begin{adjustbox}{width=0.7\textwidth,keepaspectratio}
        \begin{tabular}{c|c|c|c|c|c|c|c|c|c|c|c|c}
            \hline
            {Models} & \multicolumn{4}{|c|}{Sexist} & \multicolumn{4}{c|}{Anti-sexist} & \multicolumn{4}{|c}{Neither}\\[0.1mm]
            \cmidrule{2-13}
             {} & {R} & {C} & {Z} & {F} & {R} & {C} & {Z} & {F} & {R} & {C} & {Z} & {F}\\
             \hline
             {Mistral} & 5.00 & 4.99 & 5.00 & 5.00 & 5.00 & 4.99 & 5.00 & 5.00 & 5.00 & 4.99 & 4.99 & 5.00 \\
            {Llama} & 5.00 & 4.99 & 5.00 & 5.00 & 5.00 & 4.99 & 5.00 & 5.00 & 5.00 & 4.99 & 4.99 & 5.00 \\
            {FlanT5} & - & - & - & - & - & - & - & - & - & - & - & -\\
            {Qwen} & 5.00 & 5.00 & 5.00 & 5.00 & 5.00 & 5.00 & 5.00 & 5.00 & 5.00 & 5.00 &  5.00 &  5.00\\
            {Gemma} & 5.00 & 5.00 & 5.00 & 5.00 & 5.00 & 5.00 & 5.00 & 5.00 & 5.00 & 5.00 &  5.00 &  5.00\\
            \hline
    \end{tabular}
  \end{adjustbox}  
     \caption{\small This table demonstrates the predictive entropy of the models for each class category, across the different prompt types.}
     \label{tab:avg_pred_ent}
  \break
\end{table}


\begin{table}[ht!]
\scriptsize
\centering
\setlength{\tabcolsep}{2pt}
\renewcommand{\arraystretch}{0.9}
\begin{adjustbox}{width=\textwidth,keepaspectratio}
        \begin{tabular}{cccccccccccccccc}
            \hline
            {Models} & \multicolumn{3}{c}{Mistral} & \multicolumn{3}{c}{Llama} & \multicolumn{3}{c}{FlanT5} & \multicolumn{3}{c}{Qwen} & \multicolumn{3}{c}{Gemma} \\
            \cmidrule{2-16}
             {} & {C} & {Z} & {F} &  {C} & {Z} & {F} &  {C} & {Z} & {F} &  {C} & {Z} & {F}  & {C} & {Z} & {F}\\
             \cmidrule{2-16}
             {} & \multicolumn{15}{c}{Confident (when model predictions remain same across the prompt types)}\\  
             \hline
            {R} & 0.0010 & 0.0013 & 0.0017 & 3.90e-5 & 2.88e-5 & 8.11e-6 & 0.009 & 0.009 & 0.009 & 1.85e-6 & 1.09e-6 & 1.61e-6 & 1.12e-8 & 3.54e-8 &3.76e-8\\ 
            {C} & & 3.40e-5 & 1.90e-5 & & 1.41e-5 & 3.47e-5 & & 0.007 & 0.007 & & 1.066e-6 & 1.29e-05 & & 1.89e-8 & 1.44e-8\\ 
            {Z} & & & 5.72e-5 &  & & 2.31e-5 &  & & 0.006 & & & 1.07e-6 & & &1.71e-8\\
             \hline
            {} & \multicolumn{15}{c}{Conflict (when model predictions change across the prompt types)}\\ 
            \cmidrule{2-16}
            {R} & 0.0061 & 0.0017 & 0.0008 & 4.28e-5 & 2.59e-5 & 4.37e-6 & 0.075 & 0.064 & 0.072 & 7.19e-6 & 2.12e-6 & 2.40e-6 & 3.90e-8 & 2.70e-8 &2.16e-8\\ 
            {C} & & 0.0001 & 7.79e-5 & & 1.91e-5 & 3.85e-5 & & 0.069 & 0.082 & & 6.31e-6 & 1.20e-6 & & 2.40e-8 & 2.45e-8\\ 
            {Z} & & & 3.20e-5 & & & 2.42e-5 &  & & 0.087 & & & 1.71e-6 & & &3.45e-8\\
            
            \hline
    \end{tabular}
  \end{adjustbox}  
     \caption{\small This table documents the average impact divergent score for the models, when fed with different levels of instructions. We calculate the score for both conflicts and confident predictions. Each column and row index depicted by Roleplay (\textbf{R}), Content (\textbf{C}), Zero-shot (\textbf{Z}) and Few-shot (\textbf{F}) represent each prompt type. We record the average divergence score through the whole column, with each individual row indicating the JSD score between each generated output, when shifted from one prompt type to another (row x column).}
     \label{tab:divergent_score}
  \break
\end{table}

In Table \ref{tab:avg_pred_ent}, we calculate PE\footnote{Please refer to Appendix \ref{sec:pescore} for information on the PE scores.} of each classification category over the different models and prompt types. A lower PE typically indicates greater model confidence. However, we find minimal variation in PE across categories and prompt strategies, implying that models are often uniformly confident - even when wrong. This overconfidence can be particularly problematic in content moderation settings, where certainty may falsely suggest reliability. We see almost similar level of low PE score across the models, which indicates a confident model, but does not draw significant inferences for our analysis. This suggests that the models do not differentiate well in confidence across categories—their prediction uncertainty remains constant regardless of input type or instruction level. Slight variation (e.g., 4.99) in Mistral and Llama is negligible. Overall, the models exhibit flat, non-discriminative confidence, raising concerns about their sensitivity to nuanced classification tasks. Therefore, for classification tasks, we recommend using perplexity over PE. We can conclude from our analysis that all the models demonstrate high confidence in their prediction, despite their poor performance.

\subsection{Impact of adding instructions to the model}\label{sec:imp_inst}

After estimating the confidence of each model, we inferred that the models are quite confident in their predictions. Regardless, we wanted to check if there are differences in the confidence of the models when we change between the prompt types. We use the Impact Divergent Scores ($\mathsection$\ref{sec:divergence}) to compare the log likelihood distributions between pairs of prompt types. Table \ref{tab:divergent_score} documents the average divergence scores recorded between the prompt type, exploring how model confidence varies with annotation agreement. This time, we did not add the bias tensor to illustrate how small the changes are, between the prompt types. We observe that models are more perplexed (less confident) when human annotators disagree (conflict), confirming that perplexity is sensitive to underlying label ambiguity. For instance, Mistral and Llama show marked increase in perplexity during conflict, while FlanT5 maintains a relatively flat profile. This suggests certain models are more robust to ambiguity—critical for socially sensitive classification tasks like ours.
Consistent to the box-plots in $\mathsection$\ref{sec:perplexity_score}, we see that the change between roleplay and content type of the prompt are the most prominent. In general, the divergence between roleplay and all the other prompt types are more, as compared to the divergence between the other pairs. This shows that adding content and context information could visibly influence the model's prediction, deviating them from their prior knowledge. While it may not always give a better performance ($\mathsection$\ref{sec:quant_analysis}) or a better confidence ($\mathsection$\ref{sec:validity}), adding more instructions could be a more cost-effective way to handle specific instructions for certain domain-specific tasks, when fine-tuning is not possible.

\subsection*{Qualitative Error Analysis}\label{sec:qual}

Language can be a potent vehicle for subtle sexism (and even socially acceptable), while also a driver for reinforcing equality \citep{chew2007subtly}. It is therefore always difficult to interpret anti-sexist or sexist texts when models rely heavily on surface features such as keywords \citep{dutta_2024}. \citet{gothreau_etal_2022} identify several forms of sexism in political discourse (namely hostile, benevolent, and implicit sexism) that may reflect attitudes outside of conscious awareness. Such texts often manifest through insinuations, sarcasm, humor, or cultural references, making them difficult for language models to accurately interpret. Similarly, anti-sexist expressions that resist or critique sexism are also challenging to detect, particularly when they are emotional, indirect, or embedded in critique. Misclassifying anti-sexism as sexism not only distorts public discourse but may also penalize users who challenge abuse, raising concerns about fairness and voice in automated moderation.\\

\noindent Examples of mispredictions for `\textbf{\emph{neither}}' instances:\par
\noindent{\small \textit{``I hear the French press have properly taken the measure of @MENTION. She is being referred to as The Iron Weathercock! Priceless! They have really been paying attention!''} \quad \textbf{Prediction}: both `sexist' and `anti-sexist'\par}
\noindent{\small \textit{``MENTION telling @MENTION that many of our energy problems can be solved by extracting yet more North Sea Oil. Which is curious, since Tories have spent decades telling us we cannot be independent because it is about to run out. \#PMQs''} \quad \textbf{Mispredicted as}: `sexist'\par}
\noindent e.g., {\small \textit{``@MENTION5 and who fuck cares about a bunch of weekend warriors. The only opportunity they have to wear uniforms like that is so they can take pictures like this''} \quad \textbf{Prediction}: `sexist'\par}

\vspace{0.5\baselineskip}
\noindent Examples of mispredictions for `\textbf{\emph{sexist}}' instances:\par
\noindent{\small \textit{``This excellent judge of character MENTION is about to pick the cabinet [URL]''} \quad \textbf{Prediction as}: `neither'\par}
\noindent{\small \textit{``'Here Is a nice photo for you @MENTION - who is your mate again? [URL]''} \quad \textbf{Mispredicted as}: `neither'\par}
\noindent{\small \textit{`@MENTION @MENTION @MENTION Maybe go in to the next PMQs wearing this MENTION complete with whip, lol Most of the lot opposite would drop to their knees at the familiar sight, I'm sure. I think you're amazing. Keep up the great work. [URL]"} \quad \textbf{Mispredicted as}: `neither', or even `anti-sexist' at times\par}
\noindent{\small \textit{``@MENTION @MENTION But she is the right gender.''} \quad \textbf{Prediction}: `anti-sexist' \par}

\vspace{0.5\baselineskip}
\noindent Example of mispredictions for `\textbf{\emph{anti-sexist}}' instances:\par
\noindent{\small \textit{``Many Tories are terrified of \#MENTION \&amp; that's why they've resorted to misogynist slurs''}  \quad \textbf{Prediction}: `sexist' \par}
\noindent{\small \textit{``MENTION, the editor of the Daily Mail, was MENTION's choice as Chairman of Ofcom, the UK's communications regulator. The same MENTION would ultimately have sanctioned the MENTION piece.''} \quad \textbf{Prediction}: `neither' \par}

\vspace{0.5\baselineskip}
\noindent Overall, we see that the LLMs generalize on the explicit or obvious forms of online sexism, while missing the more subtle and implicit forms, and incorrectly labeling anti-sexist texts as sexist. And yet, when provided with more instructions in the form of prompts (including examples), their performance do not necessarily improve.

\section{Discussion}

Our findings show that large language models (LLMs) frequently misclassify anti-sexist speech as sexist or neither, particularly when it is emotionally charged, sarcastic, or politically critical. These errors are not simply technical failures: they reveal political blind spots in automated systems. By mistaking resistance for harm, automated moderation risks silencing those who challenge gendered abuse, thereby reinforcing the very power dynamics it aims to disrupt. This has implications for platform governance, where reliance on automated classification shapes whose voices remain visible.

The event-based analysis shows that politically salient moments, such as harassment incidents or leadership changes, alter both the quantity and character of online speech. In such events, counter-speech often mirrors the confrontational style of the discourse it opposes, blurring distinctions for systems trained on toxic content datasets. This overlap helps explain the consistent under-recognition of anti-sexist speech, even when contextually supportive. Trigger events can catalyse spikes in hate speech, harassment, and other harmful content \citep{lupu_2023}. Understanding their effects can help platforms and researchers anticipate, detect, and mitigate abuse \citep{geddes_2023_realworld}.

Confidence measures---entropy, perplexity, and Jensen–Shannon Divergence---did not reliably indicate accuracy, and models were often overconfident when misclassifying anti-sexist speech. This underscores the limits of statistical uncertainty measures in capturing pragmatic and contextual nuance. Sensitivity to prompt design also raises concerns about reproducibility and fairness in moderation contexts.

Overall, our results point to three priorities for improving moderation practices. First, content classification frameworks must move beyond binary harmful/not-harmful schemas to recognise constructive, oppositional speech like anti-sexism, which plays a vital role in contesting online abuse. Second, moderation processes should incorporate interpretive, human-in-the-loop review, especially during politically or socially charged events that amplify harassment, so that contextual and rhetorical cues are not lost in automated processing. Third, platform policies and training resources should explicitly include examples of resistance speech, ensuring that those who challenge sexism are not inadvertently silenced. Recognising and protecting counter-speech is not just a matter of technical accuracy; it is a question of safeguarding democratic participation and equitable representation in digital spaces. By embedding sociolinguistic awareness and contextual sensitivity into moderation design, platforms can better uphold the rights of marginalised voices and foster healthier political discourse.

\section{Conclusion}

This study advances the understanding of anti-sexism as a distinct form of political counter-speech and examines the challenges of detecting it in online political discourse using large language models. By focusing on high-salience political events that shaped the tone and content of Twitter conversations, we have shown how the discursive overlap between sexist and anti-sexist language (particularly during moments of heightened controversy) leads to systematic misclassification. These patterns highlight an important theoretical point: resistant speech often mirrors the rhetorical style of the discourse it seeks to challenge, which complicates automated detection and raises questions about the sociotechnical boundaries of content moderation.

Our findings carry practical implications for both platform governance and computational social science. For social media platforms, the tendency to conflate anti-sexist counter-speech with harmful content risks silencing those who challenge gendered harassment and undermines democratic debate. Incorporating anti-sexist exemplars into training data, developing stance-aware moderation tools, and allowing for contextual review of flagged content could help address these biases. For researchers, our results suggest the need to integrate linguistic and sociopolitical theory more explicitly into model evaluation, ensuring that categories like sexism and anti-sexism are defined in ways that reflect their pragmatic and rhetorical complexity.

While our analysis used state-of-the-art LLMs, our goal was not to benchmark technical performance but to interrogate their interpretive limitations in a politically and culturally situated task. Future work could extend this approach to other forms of counter-speech, compare cross-linguistic contexts, and examine how event salience interacts with platform dynamics to shape both human and automated interpretations. By bridging conceptual framing, event-driven analysis, and model evaluation, this study contributes to ongoing debates about the social responsibilities of AI systems in managing public discourse.

\section*{Acknowledgment}
The completion of this undertaking could not have been possible without the participation and assistance of all the annotators for this project. Their contributions are sincerely appreciated and gratefully acknowledged.

\section*{Statements and Declarations}
The data for this work would be available on request from the corresponding author, due to privacy/ethical restrictions. The code for this work will be shared publicly on a Github repository, upon publication. The author(s) declared no potential conflicts of interest with respect to the research, authorship, and/or publication of this article.

\bibliography{references}
\bibliographystyle{apalike}

\appendix
\setcounter{equation}{0}
\setcounter{figure}{0}
\setcounter{table}{0}
\setcounter{page}{1}
\makeatletter
\renewcommand{\thetable}{S\arabic{table}}
\renewcommand{\thefigure}{S.\arabic{section}.\arabic{figure}}
\renewcommand{\bibnumfmt}[1]{[S#1]}
\renewcommand{\citenumfont}[1]{S#1}

\section{Data}\label{appendix:data}

\subsection{Data Statistics}\label{sec:data_stat} \hfill\break

\begin{figure}[htb!]
    \vspace*{-0.3cm}
    \centering
    \begin{minipage}[b]{.7\linewidth}
    \centering
    \includegraphics[width=\linewidth, keepaspectratio]{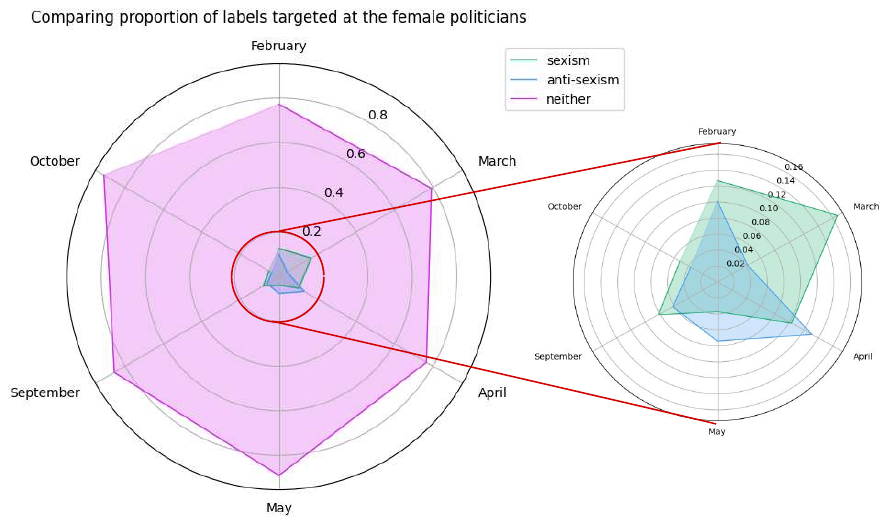}
    \end{minipage}
    \caption{\small This figure contains two radar (polar) plots comparing the proportion of three types of labels, comparing the proportion of three types of labels (\emph{sexism, anti-sexism} and \emph{neither}), targeted at female politicians across different months during periods of trigger events. We observe that though proportion of non-sexist instances are more in all the periods, the level of sexist and anti-sexist instances may co-exist in similar periods of time, indicating a possible co-dependence.}
    \label{fig:proportions_labels}
\end{figure}

\begin{table}
\centering
\footnotesize
\begin{tabular}{|c | c | c | c | c | c | c | c|} 
 \hline
 Labels per month & February & March & April & May & September & October & \textbf{Total}\\ [0.5ex] 
 \hline\hline
 Sexism & 10 & 4 & 47 & 1 & 29 & 21 & 112\\ 
 \hline
 Anti-sexism & 8 & 1 & 60 & 2 & 22 & 15 & 108\\
 \hline
 Neither & 61 & 19 & 351 & 24 & 302 & 366 & 1123\\[0.5ex]
 \hline  \hline
 \textbf{Total} & 79 & 24 & 458 & 27 & 353 & 402 & \textbf{1343}\\
 \hline

\end{tabular}
 \caption{\centering\small Data statistics showing the total number of instances for each label over the months.}
\label{tab:data_stat}

\end{table}


Although abusive content is increasingly prevalent, it still constitutes a relatively small proportion of overall posts on mainstream platforms - typically between 0.001\% and 1\%, according to prior research by academics and think tanks \citep{vidgen_2019}. Among these, sexism accounts for an even smaller subset, as online abuse encompasses various forms beyond gender-based discrimination. In our dataset of $n=1343$ (see Table~\ref{tab:data_stat}), we observe a substantially higher concentration of sexist content, particularly during periods following politically or socially charged trigger events. Figure~\ref{fig:proportions_labels} illustrates that instances of both sexism and anti-sexism tend to fluctuate in tandem, with notable peaks in March and April, suggesting a possible co-dependent relationship. This pattern implies that political discourse involving female politicians may intensify - and attract both sexist and anti-sexist responses - during high-profile events. The variation in tweet volume across months reflects the differing frequencies and public impact of the selected trigger events and was determined by ranking periods with the highest number of relevant entries. These dynamics form the basis for our comparative analysis of trigger event types in Section~\ref{sec:quant_analysis}.

\subsection{Tweet Filtering}\label{sec:filtering}

To minimize the noise, we performed the following steps in respective sequence: (1) dropping empty entries or extra spacing; (2) dropping duplicates; (3) dropping non-English texts; (4) dropping data containing only URLs or emojis (due to the vast number of emojis, one could leave their meaning based on the user's interpretation, hence they can be confusing to the classifiers); (5) remove news articles or posts mentioning the political figures through a set of keywords\footnote{This is to ensure that we focus only on the user behavior in our data, and not that of any institutions like newspapers or corporations. \textit{e.g., `BREAKING NEWS', `HEADLINES:', `In today's news', etc.}}; (6) expanding contracted texts and changing emojis to text emoticons. As the metrics of engagement, we considered sorting our data in terms of the highest number of 'retweet\_count', 'reply\_count', 'like\_count', 'quote\_count' values\footnote{These units are present in metadata of the original data, which would not be shared publicly in GitHub.} and considering the entries with highest number of engagements, with respect to the trigger events for our annotation and analysis.

\subsection{Annotation Labeling}\label{appendix:labels}
We define the three terms-- \textit{sexist}, \textit{anti-sexist}, and \textit{neither} as:

\begin{enumerate}[label=(\Alph*), leftmargin=*]
    \item \emph{Anti-sexist}: A text is anti-sexist if the speaker advocates for the equality of the sexes and the elimination of sexism and gender-based discrimination or bias-ness. It can indicate the speaker's interest to support progress towards non-violence of women and other genders, and also promoting gender equality.  In other words, anti-sexism demonstrates increasing prevalence of active behaviors challenging gender-based discrimination , i.e. anti-prejudicial sexist actions, aside from supporting reduction of the said gender-based prejudiced attitudes.  It is a commitment to challenging and combating the various forms of prejudice, bias, and inequality that can be directed toward individuals based on their gender or sex.
    \item \emph{Sexist}: A text is sexist if the speaker shows a prescriptive set of behaviors or qualities that women (and men) are supposed to exhibit in order to conform to traditional gender roles. This could be texts formulating a descriptive set of properties that supposedly differentiates the two genders, portrays women as less competent and less capable than men, and expressed through explicit or implicit comparisons and perpetuating gender-based stereotypes. 
    \item \emph{Political}: Texts revolving around discussions of politicians, policies, government actions, ideologies, elections, etc. These texts would aim to engage with societal issues, power dynamics, and decision-making processes within the realm of public affairs. Pertaining to the practice and theory of influencing other people on a civic or individual level, often concerning government or public affairs. A typical political text could have strong language, a harsh tone and slurs; and question the political standing or ideological positions of politicians or public officials.
\end{enumerate}

\subsection{Minority Voting Scheme}\label{sec:minority-label}

To address the inherent subjectivity involved in labeling complex and nuanced content such as sexism and anti-sexism, we adopt a \textit{minority voting scheme} rather than the conventional majority vote. For each tweet, we collect three labels from separate annotators, where each annotator assigns one of three possible categories: \textit{sexist}, \textit{neither}, or \textit{anti-sexist}. Instead of selecting the most frequent label, we assign the final class based on the \textit{least common (minority) vote}. This strategy is intended to preserve less dominant but potentially insightful perspectives, acknowledging that disagreement in subjective annotation tasks often reflects meaningful interpretive differences.

In the rare case where all three annotators provide different labels---a three-way tie---we apply a consistent tie-breaking rule. Notably, we did not observe any instance where one annotator labeled a tweet as sexist and another as anti-sexist, suggesting that while disagreement commonly occurs between neutrality and either pole (sexist or anti-sexist), direct contradiction between sexist and anti-sexist judgments was absent. In tie situations, our predetermined label preference follows this order: \textit{sexist} is prioritized over \textit{neither}, and \textit{neither} over \textit{anti-sexist}.

This minority-based labeling approach encourages a more inclusive and interpretive process, better capturing the nuances of online political discourse and the diversity of views regarding what constitutes sexism and its resistance.

\paragraph{Notations.}    

For each instance \( x \), labels are collected from three annotators, denoted as \( S(x) = \{s_1, s_2, s_3\} \), where each label \( s_j \in \{-1, 0, 1\} \) corresponds to one of three classes: sexist \((-1)\), neither \((0)\), or anti-sexist \((1)\), assigned by annotator \( j \). For each class \( c \in \{-1, 0, 1\} \), we calculate \( n_c \), the number of annotators assigning label \( c \).

The final label \( y(x) \) is determined by selecting the class with the \emph{lowest} frequency of votes, i.e., the minority label. This approach ensures that less common, yet meaningful perspectives are preserved, reflecting the informative nature of disagreement in subjective tasks. Formally, this can be expressed as:

\[
y(x) = \underset{c \in \{-1, 0, 1\}}{\arg\min} \, n_c \quad \text{(if no tie)}
\]

\section{Prompt Template}\label{appendix:prompt_template}

\begin{table}[ht!]
\setlength{\tabcolsep}{2pt}
\renewcommand{\arraystretch}{1} 
  \footnotesize
  \centering
  \begin{tabular}{l|p{3cm}|p{13cm}}
    \hline 
    \# & {\textbf{Prompt Category}} & \textbf{Prompt Template} (with increasing instruction on the context, content and phrasing) \\
    \hline
    1 & {Zero-shot Roleplay (R)} & {\textit{``You are an expert in understanding slight linguistic nuances in the text, even when presented with texts that lack enough context. You are well-versed with the political discourse/scenario in the United Kingdom since 2018, especially in social media platforms like Twitter.''} + 3 options to choose labels from.} \\
    2 & {Zero-shot Content (C)} & {... + Instructions for understanding `content' through linguistic cues + 3 options to choose labels from}\\
    3 & {Zero-shot Linguistic (Z)\break\citep{Radford2019LanguageMA}} & {... + Instructions for understanding `content', `context' and `phrasing' through linguistic cues + 3 options to choose labels from.}\\
    4 & {Few-shot Linguistic (F) \break \citep{brown_2020}} & {... + Instructions for understanding `content', `context' and `phrasing' through linguistic cues + 3 options to choose labels from + [representative examples \textit{(n=3)} for each label].}\\
    \hline
  \end{tabular}
  \caption{\small This table documents the different prompt categories. With each prompt from top to bottom, we increase the amount of instructions provided.}
  \label{tab:prompt categories}
\end{table}

\begin{table}[]
    \centering
    \scriptsize
    \begin{tabular}{p{1.2cm}|p{1.5cm}|p{12.8cm}}
    \hline
    \textbf{Type} & \textbf{About} & \textbf{Linguistic Information provided} \\
    \hline
    {\footnotesize Context} & {Regarding the political or non-political incident in question} & {The context is mostly political and it consists of texts that revolve around discussions of policies, government actions, ideologies, elections, etc. These texts would aim to engage with societal issues, power dynamics, and decision-making processes within the realm of public affairs. Pertaining to the practice and theory of influencing other people on a civic or individual level, often concerning government or public affairs. Reference to any of the target's current or former political and/or behavioral activity. This could be an implicit indication in the text, or a direct implication through mentions of their position on a certain topic. A typical political text could have strong language, a harsh tone and slurs; and question the political standing and political opinion of the target (usually indicated by mention of policies or government strategies) or the political position the target holds. Yet it should not undermine the intelligence of the target. 
    
    \hfill\break
    \textbf{Sexist}: Texts could be mocking female perspectives from female politicians, minimize their political contributions and undermine their achievements. It can also question their commitments to public office by implicating that they should focus more on their family commitments, and their political performance being compared to their capability in familial setting. They may also publish appearance-centric criticism of the female politicians, unlike their male counterparts. They tone could be ironic and exaggerated, and often in the guise of humor.

    \hfill\break
    \textbf{Anti-sexist}: Texts could promote female perspectives from female politicians, rebuke political gender differences, and also uplifting and encouraging more female political participation.
    
    }\\
    \hline
    {\footnotesize Content} & {Regarding what the speaker believes} & {\textbf{Sexist}: Definition as in Section \ref{sec:annotation}.
    
    \hfill\break
    \textbf{Anti-sexist}: Definition as in Section \ref{sec:annotation}.
    
    }\\
    \hline
    {\footnotesize Phrasing} & {Regarding the speaker’s choice of words} & {\textbf{Sexist}: Texts may be sexist simply because of how the speaker phrases it–independently from what general beliefs or attitudes the speaker holds. A message is sexist, for example, when it contains attacks, foul language, or derogatory depictions directed towards individuals because of their gender. However, just because a message is aggressive or uses offensive language does not mean that it is sexist.
    
    \hfill\break
    \textbf{Anti-sexist}: Texts could include reprimanding sexism both at a structural and institutional level; actively opposing the spread and occurrence of events/discussions \
    leading to sexism through challenging and confronting sexist behaviors; or attitudes of someone through a direct or a quoted text. Such texts could contain strong words in favor of demolishing institutional and structural practices of sexism and gender inequality; and may attack or rebuke targeted texts (e.g. quotes) or individuals for their sexist remarks. Texts of this kind can promote equal treatment with respect to gender and gender equality. It can discuss a gender-specific impact without expressing bias. 
    
    }\\
    \hline
  \end{tabular} 
  \caption{The prompts are further elaborated based on the linguistic cues of each instruction types.}
  \label{tab:def_terms}
  \vspace{-20pt}
\end{table}

\section{Model Uncertainty and Confidence Measures}
\label{sec:uncertainty}

\subsection{Confidence Estimation}\label{sec:confidence_estimation}\quad

Confidence (or uncertainty) estimation is crucial for learning the level of assurance or certainty associated with a model prediction, and it may acquire task-relevant confidence \citep{geng_2024}.
Measuring uncertainty in automated tasks remain as one of the most important challenges, and would be a good development in the study of political discourse \citep{Grimmer_Stewart_2013}.
The total uncertainty of a model prediction can be understood using the predictive entropy (PE) of their output distribution, encompassing both epistemic and aleatoric uncertainty \citep{Kuhn_2023}. Usually, an LLM uncertainty is measured using log probability of each generated token, where lower uncertainty indicates higher output quality, thereby providing insights on the model's confidence in its responses \citep{parthasarathy2024}. \citet{malinin_2021} state that token-level uncertainty for auto-regressive models like the LLMs is synonymous to unstructured certainty, hence they opt for a sequence-level uncertainty as that have stronger conditional independence assumptions.  

The total uncertainty of a prediction, also can be called the predictive entropy of the output distribution, measures the information one has about the output given the input \citep{Kuhn_2023}. Although it is said to be the highest when the output is minimally informative (prediction of same probability for all possible outcomes), classification tasks tend to be have lesser information than natural language generation tasks. But we follow the same three steps \citet{Kuhn_2023} follows: generation, clustering and entropy estimation.

\subsubsection*{Step 1: Generating a set of answers from the model}\quad

Given any input $x_i$, which contain the prompt $P$ and the classification instruction $q_i$, along with the varied levels of context $c_i$, we sample $K$ model outputs $Y = [y_{i, 1}, \dots, y_{i, k}, \dots, y_{i, K}]$ based on multiple iterations of the model with different seed values. For each $x_i$ containing our prompt at any level of instruction, it produces an output $Y \in \{Y_1, Y_2, Y_3\}$ corresponding to the categories of classification. 
We then group the outputs according to their classification results, into $V$ groups $G = [g_1, \dots, g_v, \dots, g_V]$, where $1 \leq V \leq T$ where $T$ is the maximum number of groups formed and would be equal to the number of classification categories we have (here, $T=3$). The uncertainty is then estimated by the entropy between each grouped sets. Similar to \citet{malinin_2021, Kuhn_2023, marjanović_2024}, we obtain $p(g_v | x_i)$, the conditional token probabilities of the output by the model, generating the answers in $g_v$ given the input $x_i$ via Equation \ref{eq:cond_prob}.
\begin{equation}
    p(g_v | x_i) = \sum_{y_{i, k} \in g_v} p(y_{i, k} | x_i) \label{eq:cond_prob}
\end{equation}

\subsubsection*{Step 2: Clustering (if needed) by semantic equivalence}\quad

In case the models do not generate outputs with the exact classification categories $c$, we would find the most semantically equivalent category $c'$ that the output corresponds to, which could be in the form of sequences. A category $c$ means the same thing as a second sequence $c'$, if and only if they entail (i.e., logically imply) each other, hence $c'$ textually entails $c$. E.g., The category `sexist' entails `I think the speaker shows slight sexist attitude', because they mean the same thing. Semantic equivalence relation, denoted by $E(\cdot, \cdot)$ induces semantic equivalence groups that are sets of categories or sequences which share the same meaning. For the space of semantically equivalent groups the categories and sequences in the set $v \in V$, all share a meaning such that $\forall c, c' \in v: E(c, c')$. 

Given that LLMs exhibit some randomness and creativity in their outputs based on prompt engineering, it makes consistent evaluation difficult in a classification tasks. Generally there are various metrics in natural language inference (NLI \citep{nli}) which are used for calculating the semantic similarity\footnote{Semantic similarity measures the similarity between two texts based on their meaning.}, such as ROUGE \citep{rouge} and BERTScore \citep{zhang2020bertscoreevaluatingtextgeneration} to evaluate semantic similarity of the generated sequence (or hypothesis sentence) with reference sequence (premise text, which would be the class categories in our case) and provide token-level granularity for recognizing the textual entailment. Yet, in our work, we realized through experimenting that often the generated output would \emph{not} contain the similar tokens or words as the reference categories. Therefore, to calculate semantic similarity between the generated output (in case the output does not match any of the classification categories) to one of the classification categories, we use DeBERTA Natural Language Inference (NLI) model\footnote{https://huggingface.co/microsoft/deberta-large-mnli} \citep{deberta}, using pairwise entailment relationships between each output instance within the output pool and the three categories for this study.
\noindent\paragraph{Implementation Details.} Following prompt engineering, we first perform some general post-processing of the text, such as -- lowering the output text, removing extra spaces or new lines, removing remnants of the chat format from the text, etc. Following which, we used the DeBERTA model for the remaining texts needed textual entailment. As inputs, we provide one output instance from the prompting ($a_i$) with each of the class categories, constructed as ${Y_i}_{i=1, 2, 3}\text{ }[SEP]\text{ }[CLS]_{a_i}$. To ensure that the output entails any of the class categories, we train it in a contrastive manner to distinguish the similarity between the two input texts, having different class category each time (output remains the same). The prompt outputs are regarded as semantically similar to one of the classes when both the inputs are in entailment, i.e., to the class category the output ($a_i$) matches the most. Furthermore, during post processing step, some models produced sparse random words, which were removed or replace with 'neither' category. The AMCE emphasizes that the estimand itself is conditional on, and therefore depends on, the distribution of the other candidate attributes included in the study.

After all the output sequences represent any of the class categories, we calculate the conditional distributions over tokens and their resulting sequences, the probability of the sequence conditioned on the context level all the previous $y_{1:i-1}$, comes from the conditional token probabilities $p(y|x) = \Pi_i p(y_i | y_{1:i-1}, x)$. This is because auto-regressive models are used to factorize the joint distribution over $y$ into a product of conditionals over a finite set of tokens. And this set of conditional independence assumptions allows us to define a model on the finite space $\mathcal{Y}_L$, where $L$ is the number of tokens in a sequence. 
Since we are focused on probability of the model generating categories or sequences that share the same meaning, Equation \ref{eq:cond_prob} can be modified to:
\begin{equation}
    p(g_v | x_i) = \sum_{y_{i, k} \in g_v} p(y_{i, k} | x_i) = \sum_{y_{i, k} \in g_v} \mathlarger{\mathlarger{\mathlarger{\Pi}}}_{i}  p(y_{i, k} | y_{i-1, k}, x_i) \label{eq:cond_prob_mod}
\end{equation}

\subsubsection*{Step 3: Computing predictive entropy}\quad

Post the determination of each cluster of generated categories or sequences that mean the same thing, we add their likelihoods following Equation \ref{eq:cond_prob_mod} as a way of determining the likelihood of each meaning rather than each sequence. The PE, i.e., the total uncertainty \citep{depeweg_2018} related to the prediction $y_q$ for a concrete query instance $x_q \in \mathcal{X}$, is given by the entropy of the predictive posterior, which is given by taking the expectation over the ensemble.
In other words, using PE, we are wondering what can be said about the (ambiguous) nature of the text instance, especially about the uncertainty related to the prediction of that outcome. Thus, PE is computed as the entropy over meaning-distribution,
\begin{equation}
    PE(x_i) = \mathbb{E}_{y_{i, k} \in g_v} [ - \log p(y_{i, k} | x_i)] = - \sum_v p(g_v | x_i) \log p(g_v | x_i) 
    = - \sum_v \mathlarger{\mathlarger{(}}\mathlarger{(}\sum_{y_{i, k} \in g_v} p(y_{i, k} | x_i)\mathlarger{)}\log [\sum_{y_{i, k} \in g_v} p(y_{i, k} | x_i)]\mathlarger{\mathlarger{)}} \label{eq:p_ent}
\end{equation}

One of the challenges in auto-regressive models is that all expressions till Equation \ref{eq:p_ent} are intractable to evaluate, and therefore would need Monte-Carlo approximation to make it tractable \citep{malinin_2021}. 
With the probabilities of different groups $V$, we approximate the overall expectation of predictive entropy $PE(x_i)$ using Monte Carlo integration over the groups:
\begin{equation}
    PE(x_i) \approx - |V|^{-1} \sum_{v=1}^{|V|} \log p(g_v | x_i) \label{eq:monte-carlo}
\end{equation}

\paragraph{Understanding the PE scores.}\label{sec:pescore}
When calculating entropy in NLP tasks and dealing with logarithmic values and probabilities (like in our study), a bias tensor is typically added to ensure numerical stability. As we see in Equation \ref{eq:p_ent}, if we consider the whole probability portion as $p_i$, we can re-write the formula as $PE(x_i) = -\sum p_i \log p_i$. Hypothetically, if a model predicts something with $0$ probability, it would make the entropy undefined (as $\log(0) = -\infty$) which would break the training process. Therefore, we add a small bias tensor (i.e., a constant to each probability before computing the logarithm, denoted by $\epsilon$) to avoid the aforementioned issues. Thus, the modified formula can be written as $PE(x_i) = -\sum p_i \log (p_i + \epsilon)$. Following \citet{Kuhn_2023}, we added a constant 5, which explains why most of the values are approximately 5. Yet, lesser the value, the better, as that indicates lesser uncertainty of the model. 

\subsection{Perplexity Measure}\label{sec:perplexity}\quad

Perplexity is a metric commonly used in NLP to evaluate the quality of the LLMs, particularly in the context of text generation. Much like confidence estimation, it measures the "surprise" of a language model in predicting the next word in a sequence of words, given the set of instructions (or context), which could be in the form of prompts.
Therefore, a high perplexity indicates that the LLM is not confident in its text generation — that is, the model is "perplexed" — whereas a low perplexity indicates that the language model is confident in its generation. 
In other words, perplexity is a measurement of uncertainty in the predictions of a language model. Mathematically, it is defined as the exponential of the average negative log-likelihood of a sequence of tokens.

\begin{equation}
    \text{PPL} = \exp\mathlarger{(}{-\frac{1}{L}\sum_i^L p(y_i|y_{i - 1}, x_i, c_i)}\mathlarger{)} \label{eq:perplexity}
\end{equation}
where $y_i$ represents the token that is generated and $y_{i-1}$ represents the previous token, while $x_i$ and $c_i$ indicate the text and the context information (provided as prompts) respectively -- all three oof them influence the condition. 

\subsection{Impact Divergent Scores}\label{sec:divergence}\quad

In-context learning (ICL) in the form of instructing the LLMs through prompting have shown to elevate models' performance in text classification across different domains, without having to rely on fine-tuning \citep{edwards_2024}. But does adding more instruction (zero-shot) with examples (few-shot) really improve the models' performance, or do they retain the knowledge learnt from the pre-trained data. We analyze this through a novel method called \emph{impact divergent score} across the different levels of instructions we provide the model.
We utilize divergence scores to gauge the log likelihood attributed to the generated answer in the output distribution (given the input $x_i$ and prompt $c_i$). Since ours is a classification task with all the outputs representing either of the classes, we opted for a divergence score as used in \citet{dutta_2024a}, called Jensen–Shannon divergence (JSD) -- a method for measuring the similarity between two probability distributions. It is a symmetric and smoothed version of the Kullback–Leibler divergence $KL(P||Q)$ denoted by $\sum_i P(i)*\log(P(i)/ Q(i))$, where $P$ and $Q$ are the target and predicted probability distributions respectively. It is denoted as $JSD(P||Q) = \frac{1}{2}KL(P||M) + \frac{1}{2}KL(Q||M)$ where the value of $M$ is calculated as the average of $P$ and $Q$, i.e., $M = (P + Q)/2$. The score for each instance is calculated as the divergent distance between the averaged softmax probability distribution of all the tokens in the generated output, from one prompt type to another.
A JSD score of $0$ would indicate two identical distributions, while a score closer to $1$ would indicate very different distributions. Any intermediate score would indicate the degree of similarity or dissimilarity between the two distributions. Given the distance between the probability distributions between pairs of $y_{i;r=R,q}$, $y_{i;r=C,q}$, $y_{i;r=Z,q}$ and $y_{i;r=F,q}$, it will indicate if the LLM’s probability distribution is swayed by the given instructions, while ensuring that small changes in syntactic representation or mismatches from first-token probabilities are not included.

\section{Extended Figures and Results}
\label{appendix:figures}

\subsection{Model by model breakdown of perplexity with respective predictions}\label{appendix:perp_sex_anti}

\begin{figure}[htb!]
  \begin{minipage}[b]{\linewidth}
    \centering
    \includegraphics[width=0.9\linewidth, height=3.8cm]{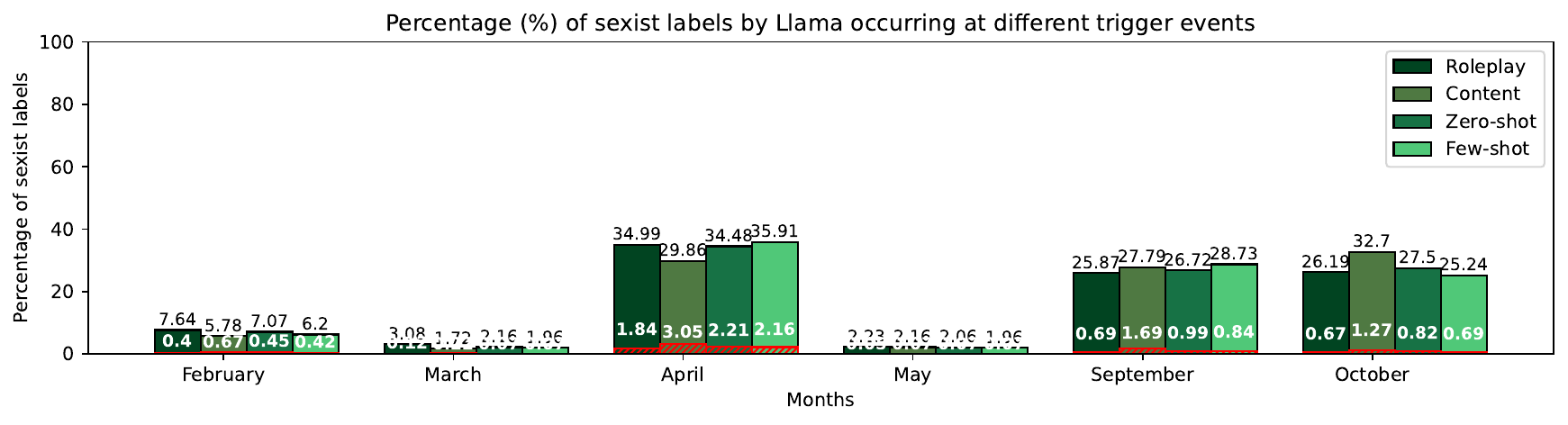}
    \captionof{figure}*{\centering\small(A): Llama}
  \end{minipage}
  \break
  \begin{minipage}[b]{\linewidth}
    \centering
    \includegraphics[width=0.9\linewidth, height=3.8cm]{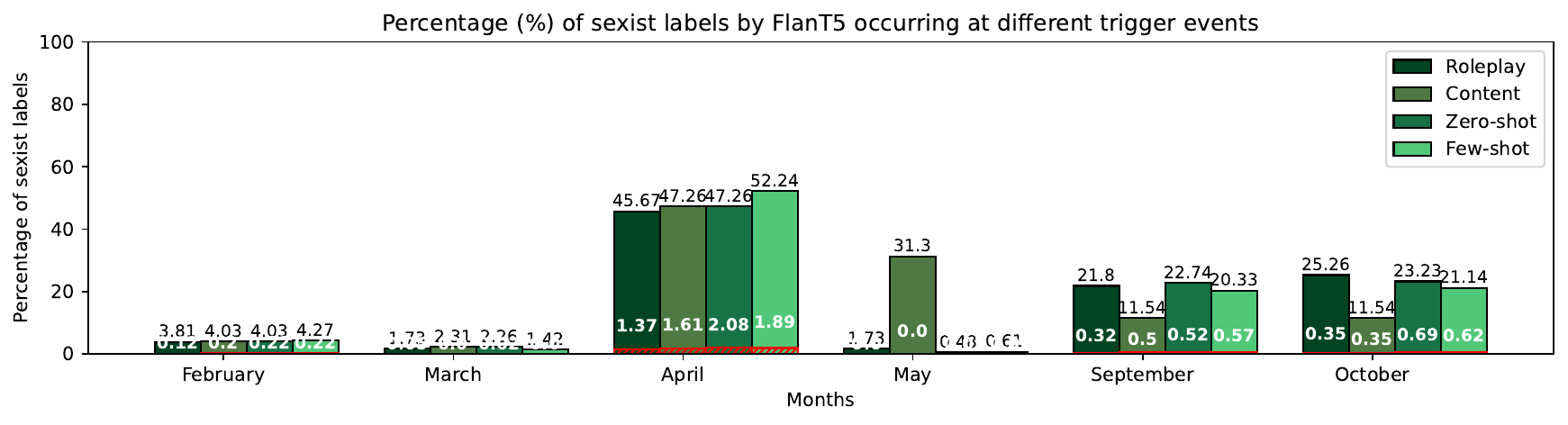}
    \captionof{figure}*{\centering\small(B): FlanT5}
  \end{minipage}
  \break
  \begin{minipage}[b]{\linewidth}
    \centering
    \includegraphics[width=0.9\linewidth, height=3.8cm]{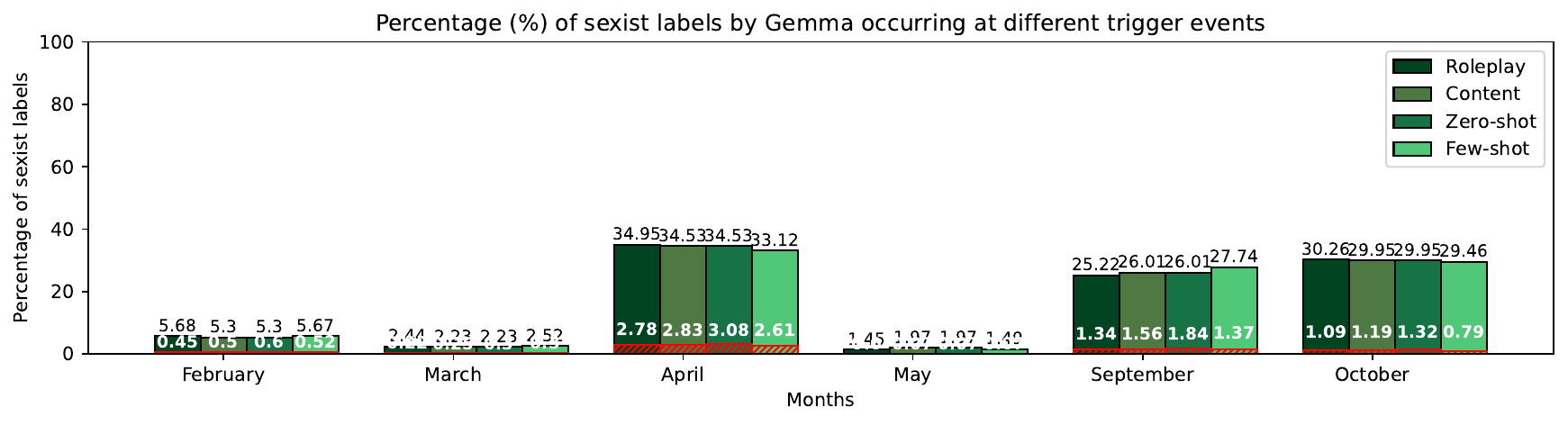}
    \captionof{figure}*{\centering\small(C): Gemma}
  \end{minipage}\hfill
  \caption{\small The figures showcase bar plots representing the proportion of predictions that were detected as sexist by the models in the given month, with respect to each prompt type. We also record the proportion of those predictions that were correctly predicted, represented as red shaded bars for each month and prompt type.}
\label{fig:sexism_prop}
\end{figure} 

\begin{figure}[htb!]
  \begin{minipage}[b]{\linewidth}
    \centering
    \includegraphics[width=0.9\linewidth, height=3.8cm]{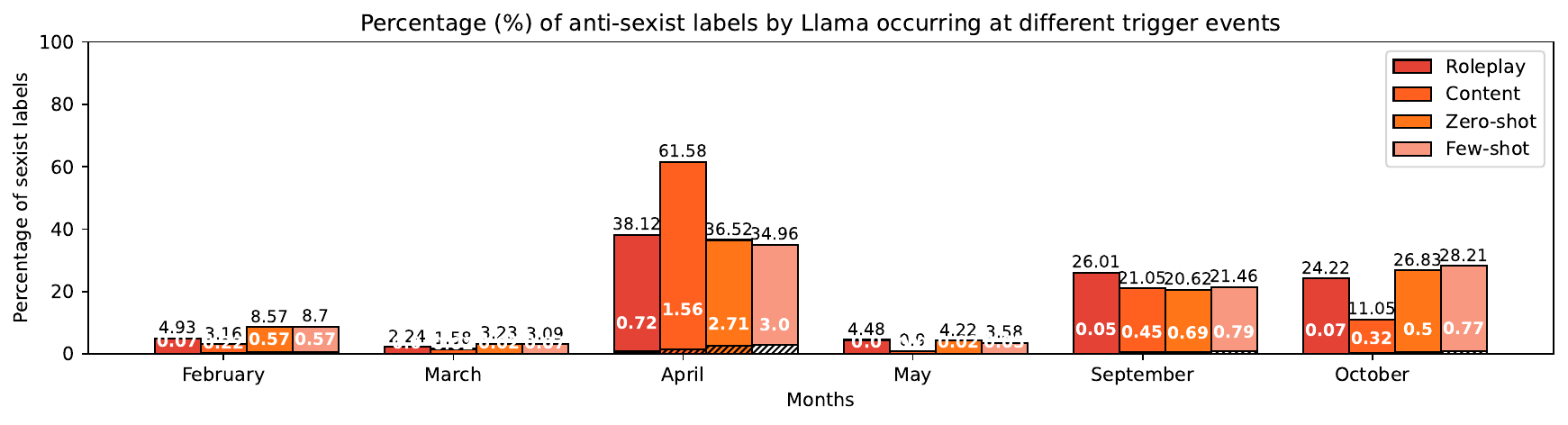}
    \captionof{figure}*{\centering\small(A): Llama}
  \end{minipage}
  \break
  \begin{minipage}[b]{\linewidth}
    \centering
    \includegraphics[width=0.9\linewidth, height=3.8cm]{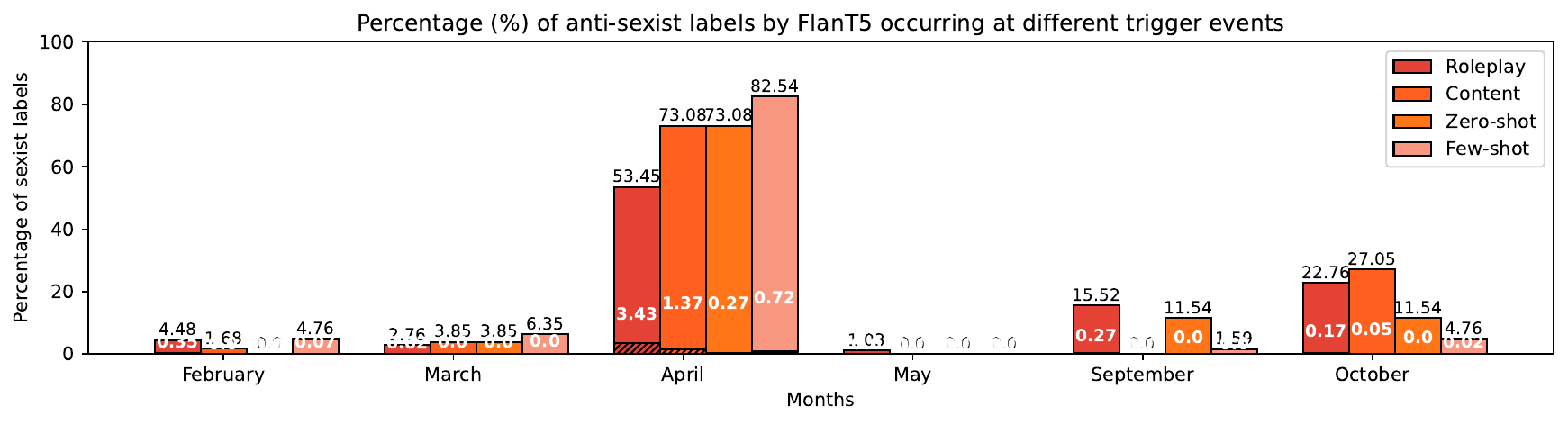}
    \captionof{figure}*{\centering\small(B): FlanT5}
  \end{minipage}
  \break
  \begin{minipage}[b]{\linewidth}
    \centering
    \includegraphics[width=0.9\linewidth, height=3.8cm]{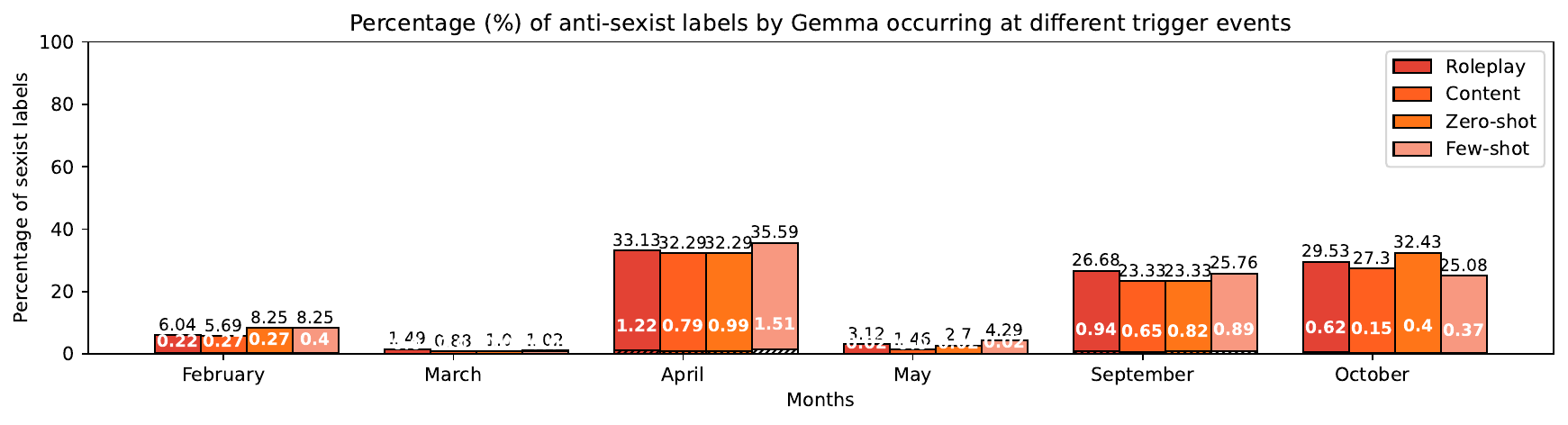}
    \captionof{figure}*{\centering\small(C): Gemma}
  \end{minipage}\hfill
  \caption{\small The figures showcase bar plots representing the proportion of predictions that were detected as anti-sexist by the models in the given month, with respect to each prompt type. We also record the proportion of those predictions that were correctly predicted, represented as black shaded bars for each month and prompt type.}
\label{fig:antisexism_prop}
\end{figure} 

\subsection{Predicted label distribution by model across the full dataset}\label{appendix:pred_label_mod}

\begin{figure}[htb!]
    \centering   
    \begin{minipage}[b]{.9\linewidth}
    \centering
    \includegraphics[width=\linewidth, keepaspectratio]{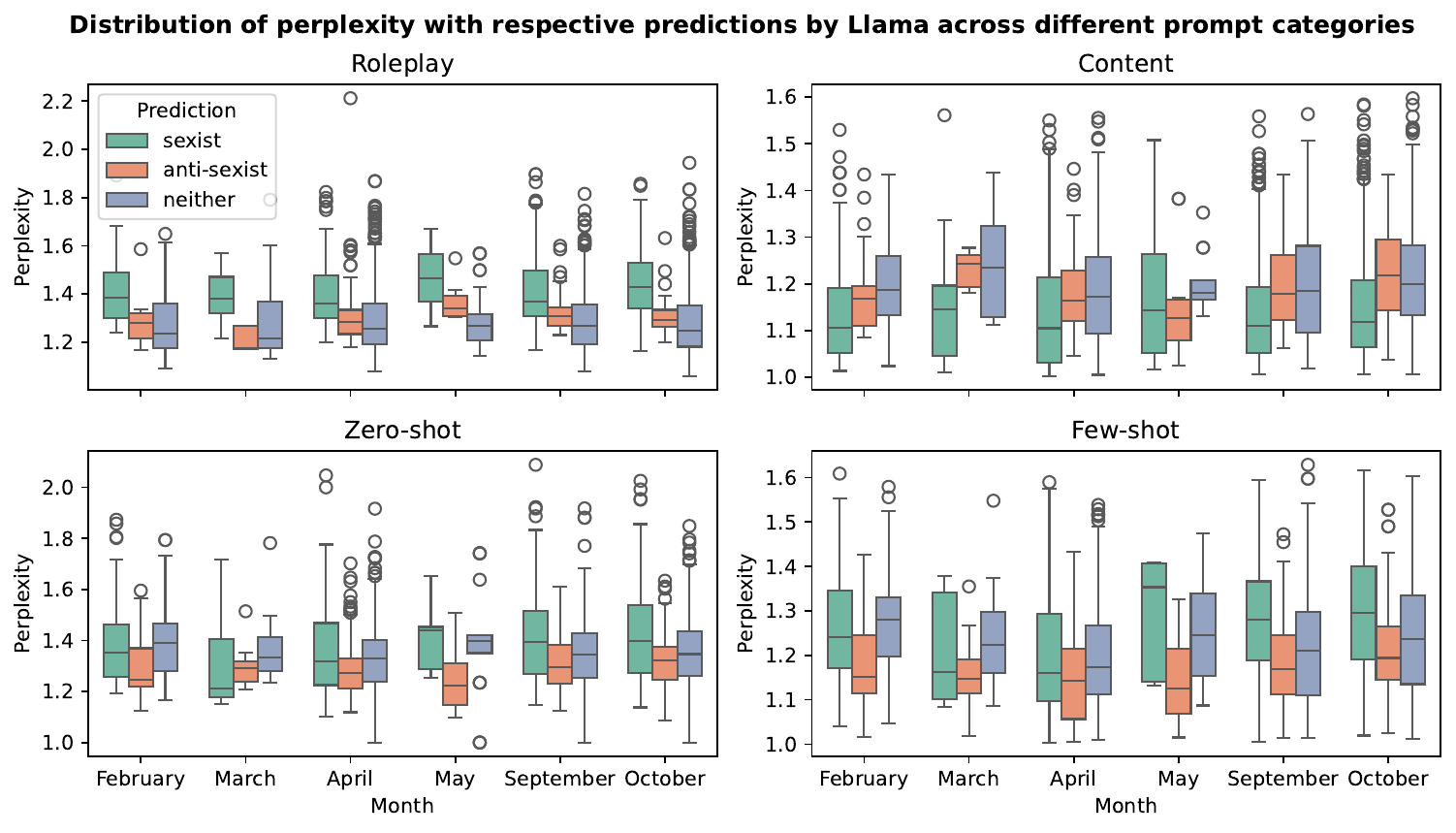}
    \captionof{figure}*{\centering\small(A): }
  \end{minipage}\hfill
  \break
  \begin{minipage}[b]{.7\linewidth}
    \centering
    \includegraphics[width=\linewidth, keepaspectratio]{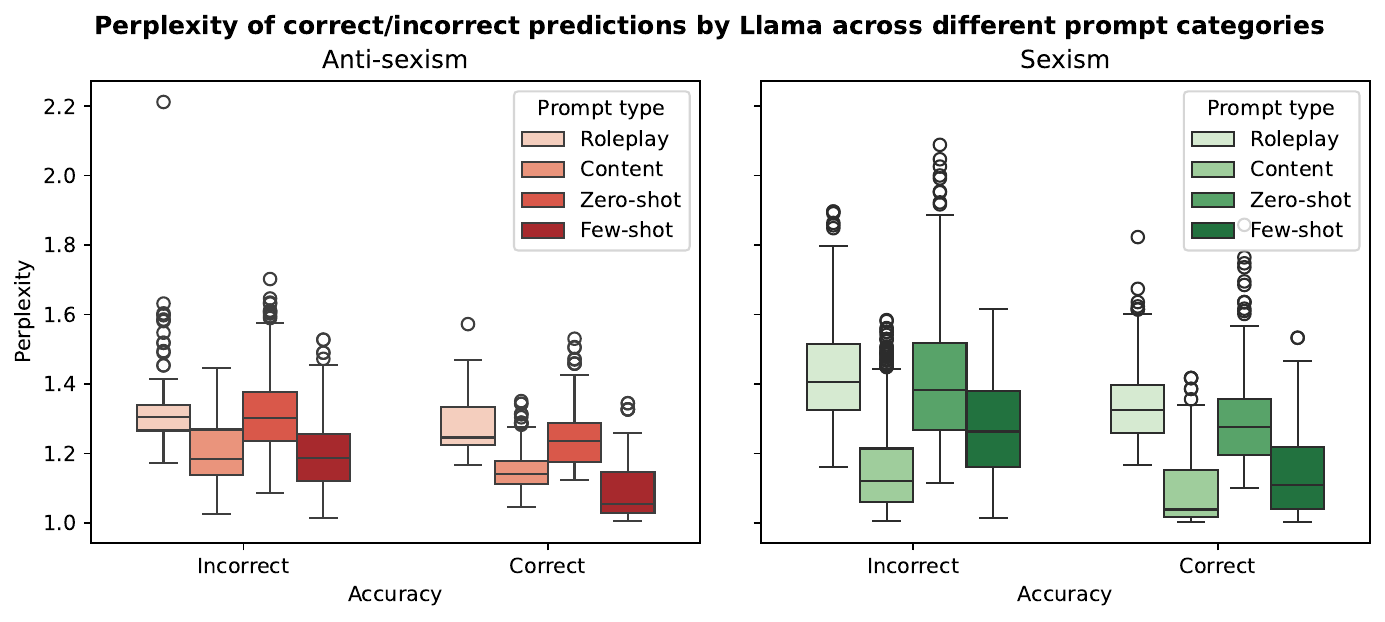}
    \captionof{figure}*{\centering\small(B): }
  \end{minipage}\hfill
  \caption{\small The figures demonstrate boxplots representing the distribution of the perplexity score for Llama.}
\label{fig:llama_perp}
\end{figure} 

\begin{figure}[htb!]
    \centering   
    \begin{minipage}[b]{\linewidth}
    \centering
    \includegraphics[width=\linewidth, keepaspectratio]{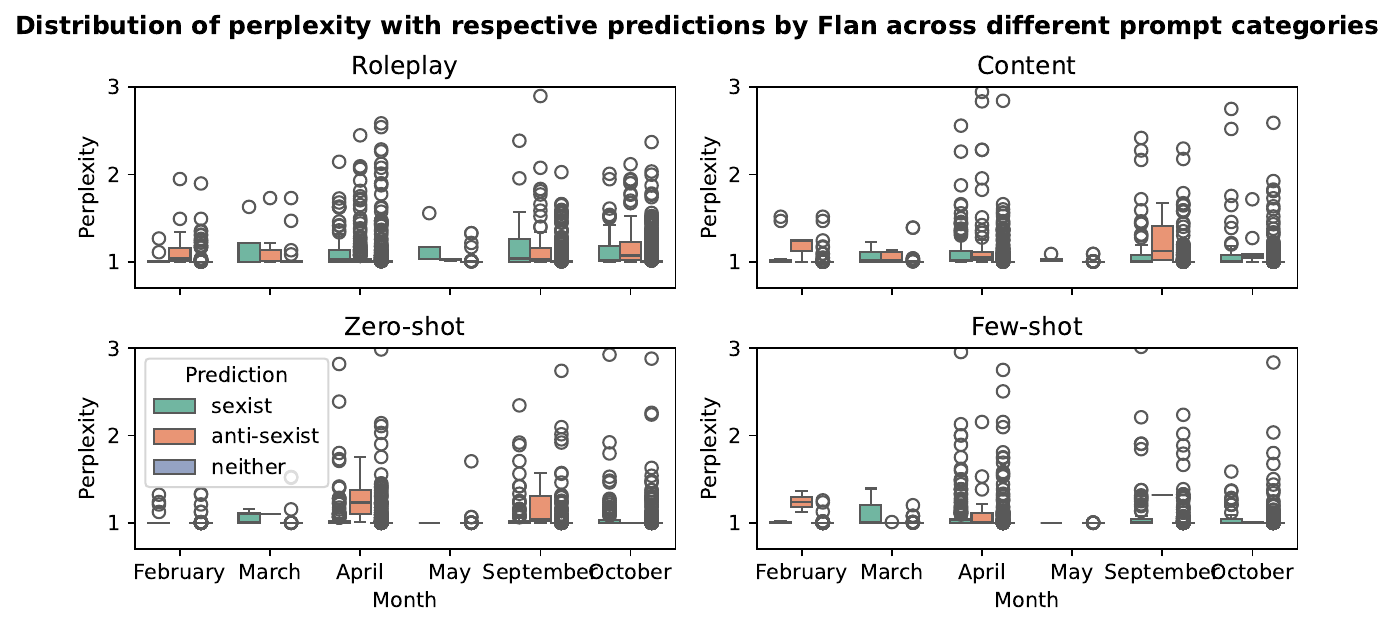}
    \captionof{figure}*{\centering\small(A)}
  \end{minipage}\hfill
  \break
  \begin{minipage}[b]{.7\linewidth}
    \centering
    \includegraphics[width=\linewidth, keepaspectratio]{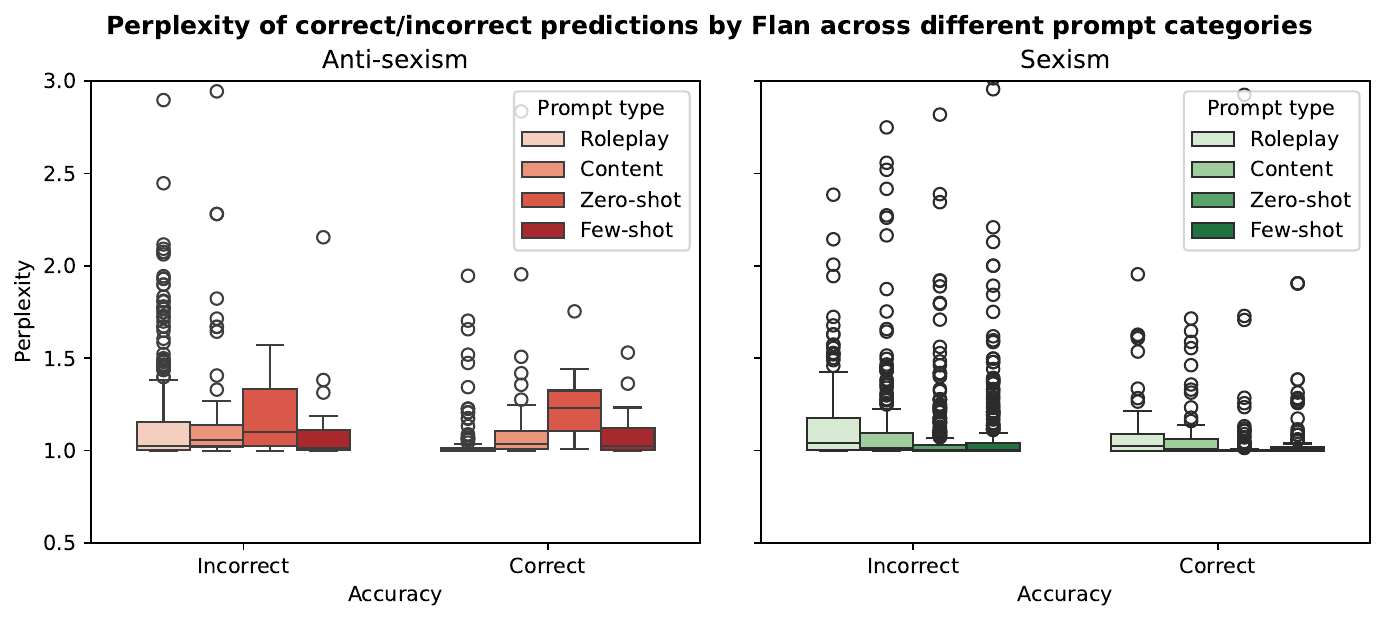}
    \captionof{figure}*{\centering\small(B)}
  \end{minipage}\hfill
  \caption{\small The figures demonstrate boxplots representing the distribution of the perplexity score for FlanT5.}
\label{fig:flan_perp}
\end{figure} 

\begin{figure}[htb!]
    \centering   
    \begin{minipage}[b]{.9\linewidth}
    \centering
    \includegraphics[width=\linewidth, keepaspectratio]{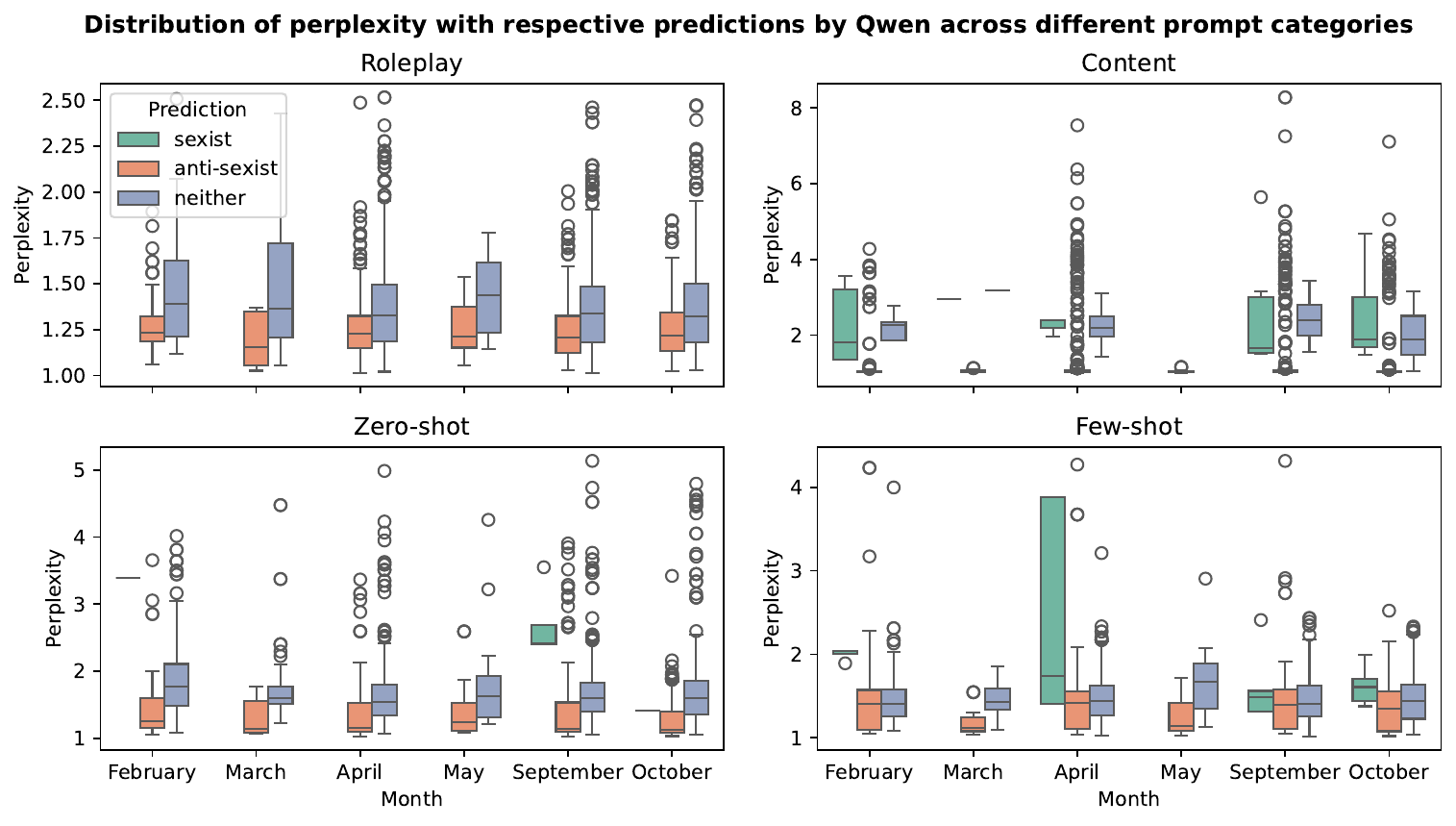}
    \captionof{figure}*{\centering\small(A)}
  \end{minipage}\hfill
  \break
  \begin{minipage}[b]{.7\linewidth}
    \centering
    \includegraphics[width=\linewidth, keepaspectratio]{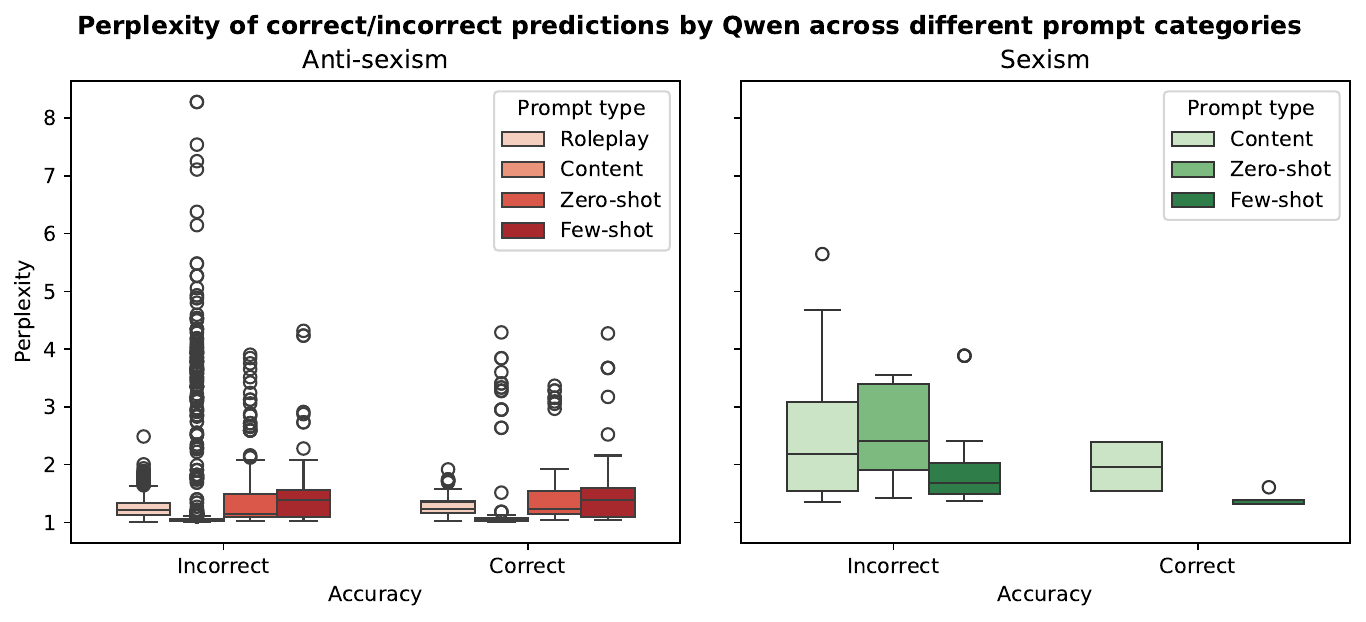}
    \captionof{figure}*{\centering\small(B)}
  \end{minipage}\hfill
  \caption{\small The figures demonstrate boxplots representing the distribution of the perplexity score for Qwen.}
\label{fig:qwen_perp}
\end{figure} 

\begin{figure}[htb!]
    \centering   
    \begin{minipage}[b]{.9\linewidth}
    \centering
    \includegraphics[width=\linewidth, keepaspectratio]{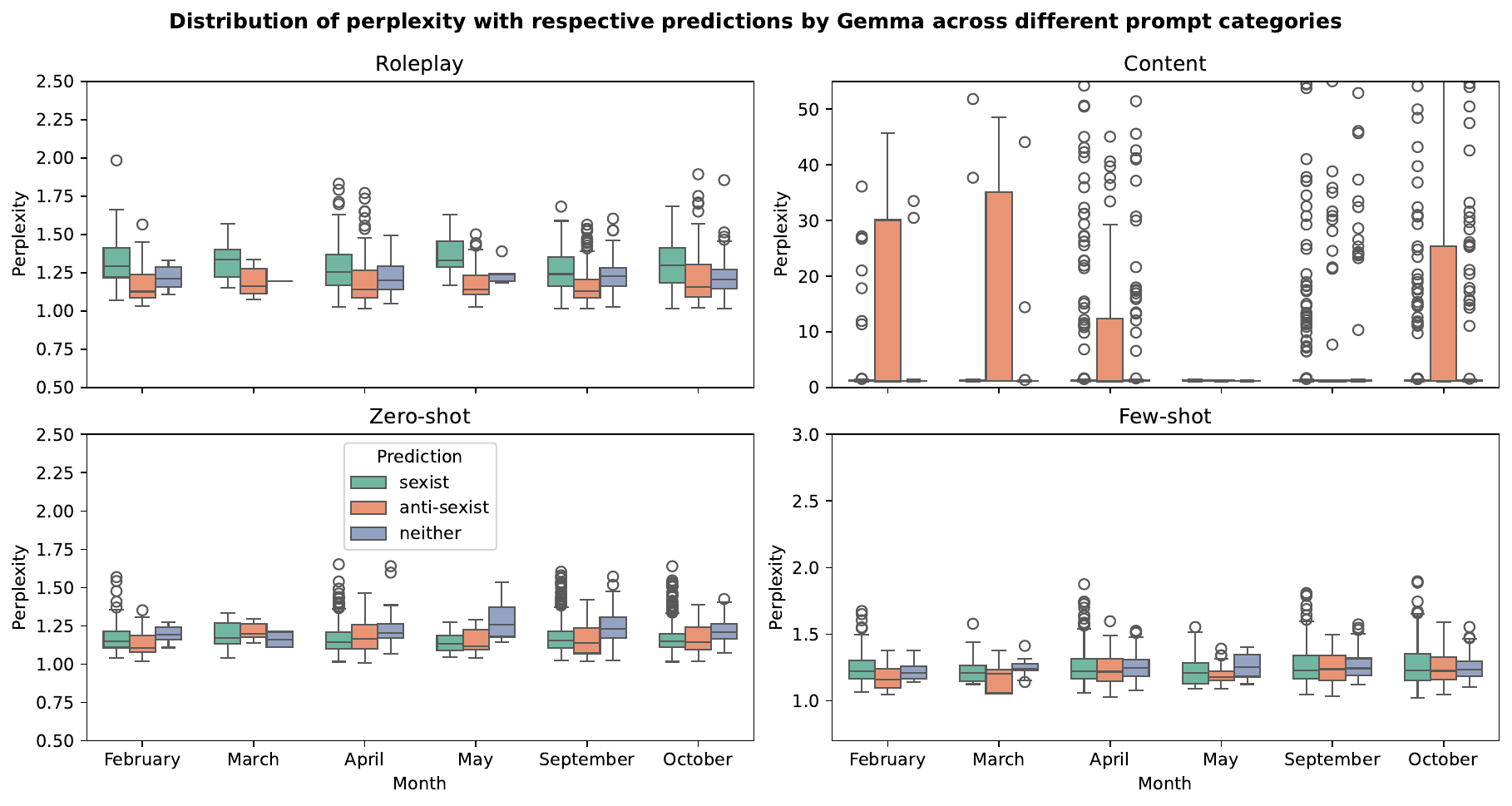}
    \captionof{figure}*{\centering\small(A)}
  \end{minipage}\hfill
  \break
  \begin{minipage}[b]{.7\linewidth}
    \centering
    \includegraphics[width=\linewidth, keepaspectratio]{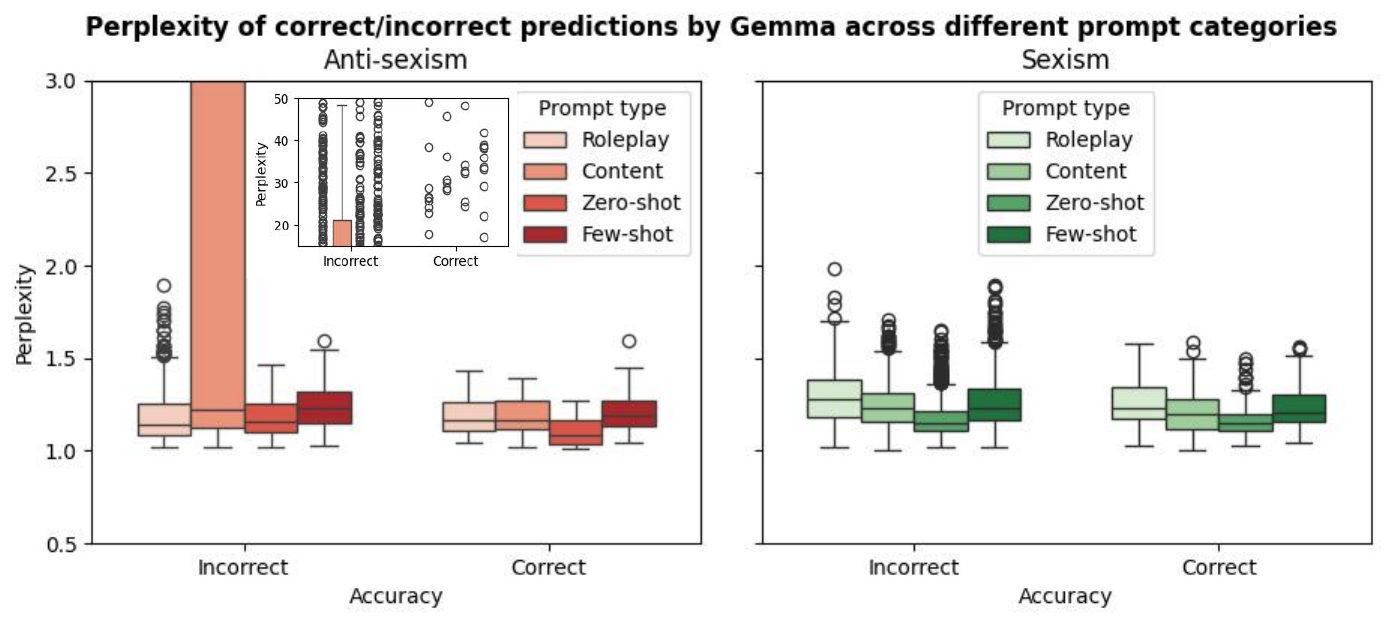}
    \captionof{figure}*{\centering\small(B)}
  \end{minipage}\hfill
  \caption{\small The figures demonstrate boxplots representing the distribution of the perplexity score for Gemma.}
\label{fig:gemma_perp}
\end{figure}

\end{document}